\documentclass[sigconf, screen]{acmart}
\AtBeginDocument{%
  \providecommand\BibTeX{{%
    \normalfont B\kern-0.5em{\scshape i\kern-0.25em b}\kern-0.8em\TeX}}}

\usepackage{cleveref}
\usepackage{subcaption}
\usepackage{enumitem}

\copyrightyear{2022}
\acmYear{2022}
\setcopyright{rightsretained}
\acmConference[KDD '22]{Proceedings of the 28th ACM SIGKDD Conference on Knowledge Discovery and Data Mining}{August 14--18, 2022}{Washington, DC, USA}
\acmBooktitle{Proceedings of the 28th ACM SIGKDD Conference on Knowledge Discovery and Data Mining (KDD '22), August 14--18, 2022, Washington, DC, USA}
\acmDOI{10.1145/3534678.3539301}
\acmISBN{978-1-4503-9385-0/22/08}

\usepackage{etoolbox}
\makeatletter
\patchcmd{\maketitle}{\@copyrightpermission}{
   \begin{minipage}{0.3\columnwidth}
     \href{https://creativecommons.org/licenses/by/4.0/}{\includegraphics[width=0.90\textwidth]{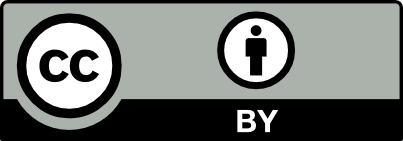}}
   \end{minipage}\hfill
   \begin{minipage}{0.7\columnwidth}
     \href{https://creativecommons.org/licenses/by/4.0/}{This work is licensed under a Creative Commons Attribution International 4.0 License.}
   \end{minipage}

   \vspace{5pt}
}{}{}

\makeatother

\begin{document}

\title{Deep Representations for Time-varying Brain Datasets}

\author{Sikun Lin}
\affiliation{%
  \institution{University of California, Santa Barbara}
  \city{Santa Barbara}
  \state{CA}
  \country{USA}
}
\email{sikun@ucsb.edu}

\author{Shuyun Tang}
\affiliation{%
  \institution{University of California, Santa Barbara}
  \city{Santa Barbara}
  \state{CA}
  \country{USA}
}
\email{shuyun@ucsb.edu}

\author{Scott T. Grafton}
\affiliation{%
  \institution{University of California, Santa Barbara}
  \city{Santa Barbara}
  \state{CA}
  \country{USA}
}
\email{scott.grafton@psych.ucsb.edu}

\author{Ambuj K. Singh}
\affiliation{%
  \institution{University of California, Santa Barbara}
  \city{Santa Barbara}
  \state{CA}
  \country{USA}
}
\email{ambuj@cs.ucsb.edu}

\renewcommand{\shortauthors}{Sikun Lin, et al.}

\begin{abstract}
Finding an appropriate representation of dynamic activities in the brain is crucial for many downstream applications. Due to its highly dynamic nature, temporally averaged fMRI (functional magnetic resonance imaging) can only provide a narrow view of underlying brain activities. Previous works lack the ability to learn and interpret the latent dynamics in brain architectures. This paper builds an efficient graph neural network model that incorporates both region-mapped fMRI sequences and structural connectivities obtained from DWI (diffusion-weighted imaging) as inputs. We find good representations of the latent brain dynamics through learning sample-level adaptive adjacency matrices and performing a novel multi-resolution inner cluster smoothing. 
We also attribute inputs with integrated gradients, which enables us to infer (1) highly involved brain connections and subnetworks for each task, (2) temporal keyframes of imaging sequences that characterize tasks, and (3) subnetworks that discriminate between individual subjects. This ability to identify critical subnetworks that characterize signal states across heterogeneous tasks and individuals is of great importance to neuroscience and other scientific domains. Extensive experiments and ablation studies demonstrate our proposed method's superiority and efficiency in spatial-temporal graph signal modeling with insightful interpretations of brain dynamics.
\end{abstract}





\maketitle

\section{Introduction}
Neuroimaging techniques such as fMRI (functional magnetic resonance imaging) and DWI (diffusion-weighted imaging) provide a window into complex brain processes. Yet, modeling and understanding these signals has always been a challenge. Network neuroscience \citep{bassett2017network} views the brain as a multiscale networked system and models these signals in their graph representations: nodes represent brain ROIs (regions of interest), and edges represent either structural or functional connections between pairs of regions.

With larger imaging datasets and developments in graph neural networks,
recent works leverage variants of graph deep learning, modeling brain signals with data-driven models and getting rid of Gaussian assumptions that typically existed in linear models \citep{zhang2019estimating,li2019modeling}.
These methods are making progress on identifying physiological characteristics and brain disorders:
In \cite{10.3389/fnins.2020.00630}, authors combine grad-CAM \citep{selvaraju2017grad} and GIN \citep{xu2018powerful} to highlight brain regions that are responsible for gender classification with resting-state fMRI data. Others \cite{Li2020PoolingRG} propose to use regularized pooling with GNN to identify fMRI biomarkers. However, these works use time-averaged fMRI, losing rich dynamics in the temporal domain. They also do not incorporate structural modality that can provide additional connectivity information missing in the functional modality. 
Another work \cite{noman2021graph} embeds both topological structures and node signals of fMRI networks into a low-dimensional latent representations for better identification of depression, but it combines nodes' temporal and feature dimensions instead of handling them separately, leading to a suboptimal representation
(as discussed in \cref{ssc:model_comparison}).
To overcome these issues, we propose ReBraiD (Deep \textbf{Re}presentations for Time-varying \textbf{Brai}n \textbf{D}atasets), a graph neural network model that jointly models dynamic functional signals and structural connectivities, leading to a more comprehensive deep representation of brain dynamics.

To simultaneously encode signals along spatial and temporal dimensions, some works in traffic prediction and activity recognition domains such as Graph WaveNet \citep{wu2019graph} alternate between TCN (temporal convolution network) \citep{10.1007/978-3-319-49409-8_7} and GCN (graph convolutional network) \citep{kipf2017semisupervised}.
Others \citep{song2020spatial,liu2020disentangling} use localized spatial-temporal graph to embed both domains' information in this extended graph. Some proposed methods also incorporate gated recurrent networks for the temporal domain such as \citep{seo2018structured,ruiz2020gated}.
We choose to alternate TCN with GCN layers for ReBraiD, as it is more memory and time-efficient and can support much longer inputs.
On top of this design, we propose novel ``sample-level adaptive adjacency matrix learning'' and ``multi-resolution inner cluster smoothing,'' both of which learn and refine latent dynamic structures. With the choice of the temporal layer, our model is more efficient than other baselines while having the highest performance.

We perform extensive ablation studies to examine individual components of the model. We also explore the best option when alternating spatial and temporal layers for encoding brain activities.
After quantitatively showing the representation ability of our model, we utilize IG (integrated gradients)  \citep{sundararajan2017axiomatic} to identify how brain ROIs participate in various processes. This can lead to better behavioral understanding, discovery of biomarkers, and characterization of individuals or groups. We also make the novel contribution of identifying temporally important frames with graph attribution techniques; this can enable more fine-grained temporal analysis around keyframes when combined with other imaging modalities such as EEG (electroencephalogram). In addition, our subject-level and group-level attribution studies unveil heterogeneities among ROIs, tasks, and individuals.

In summary, the main contributions of our work are as follows:
\begin{itemize}[leftmargin=*]
    \item We present ReBraid, an efficient graph neural network model that jointly models both structural and dynamic functional brain signals, providing a more comprehensive representation of brain activities when compared to the current fMRI literature.
    \item Unlike typical spatial-temporal GCNs that learn a universal latent structure, we propose sample-level latent adaptive adjacency matrix learning based on input snippets. This captures the evolving dynamics of a task better.
    \item We propose multi-resolution inner cluster smoothing, which effectively encodes long-range node relationships while keeping the graph structure, enabling the model to leverage structural and latent adjacency matrices throughout the process. Together with subject SC and sample-level adjacency matrix learning, the inner cluster smoothing learns and refines latent dynamic structures on limited signal data.
    \item We carry out extensive ablation studies and model comparisons to show ReBraid's superiority in representing brain dynamics. We also leverage integrated gradients to attribute and interpret the importance of both spatial brain ROIs and temporal keyframes, as well as heterogeneities among brain ROIs, tasks, and subjects. These can open up new opportunities for identifying biomarkers for different tasks or diseases and markers for other complex scientific phenomena.
\end{itemize}
\section{Method}
\label{sc:method}

\subsection{Preliminaries}
\label{sc:preliminaries}
We utilize two brain imaging modalities mapped onto a same coordinate: SC (structural connectivity) from DWI scans, and time-varying fMRI scans. We represent them as a set of $L$ graphs 
\begin{small}
$\mathcal{G}_i = (A_i, X_i)\text{ with } i\in[1,L]$.
\end{small}
\begin{small}
$A_i \in \mathbb{R}^{N \times N}$ 
\end{small} represents normalized adjacency matrix with an added self-loop:
\begin{small}
\(A_i = \tilde{D}_{\text{SC}_i}^{-\frac{1}{2}}\tilde{\text{SC}_i}\tilde{D}_{\text{SC}_i}^{-\frac{1}{2}}\), \(\tilde{\text{SC}_i}=\text{SC}_i+I_N\)
\end{small}
and
\begin{small}
\(\tilde{D}_{\text{SC}_i}=\sum_w (\tilde{\text{SC}_i})_{vw}\)
\end{small}
is the diagonal node degree matrix.
Graph signal matrix obtained from fMRI scans of the \(i^{th}\) sample is represented as \begin{small}
$X_i \in \mathbb{R}^{N\times T}$.
\end{small}
Here $N$ is the number of nodes, and each node represents a brain region; $T$ is the input signal length on each node. We refine our representation using the task of classifying brain signals $\mathcal{G}_i$ into one of $C$ task classes through learning latent graph structures.

\subsection{Model}
\label{sc:model}
\begin{figure}[t]
    \centering
    \captionsetup{font=small}
    \includegraphics[width=0.5\textwidth]{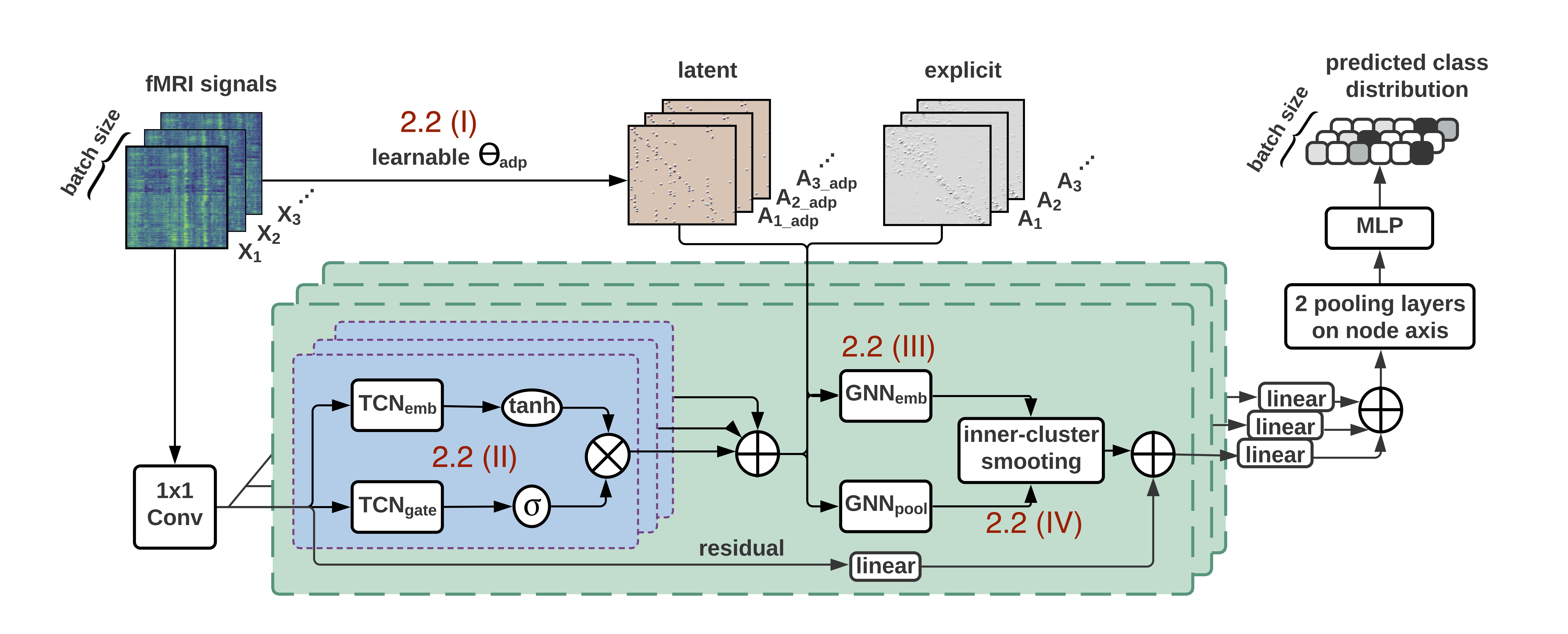}
    \caption[Caption for LOF]{The proposed ReBraiD model for integrating brain structure and dynamics (the architecture shown is for classification). For each batch with batch size $B$, input $X$ has a dimension of \begin{small}$(B, 1, N, T)$\end{small}\footnotemark, and $A, A_{\text{adp}}$ both have the dimension \begin{small}$(B, N, N)$\end{small}. The encoder (green part) encodes temporal and spatial information alternately, producing a latent representation in \begin{small}$(B, d_{\text{latent}}, N, 1)$\end{small}. These embeddings are followed by linear layers for pooling and classification. The final output has a dimension of \begin{small}$(B, C)$\end{small}.}
    \label{fig:model_structure}
\end{figure}
\footnotetext{Axis order follows PyTorch conventions. Dimension at the second index is the expanded feature dimension.}
ReBraiD takes $(A, X)$ as inputs and outputs task class predictions. The overall model structure is shown in \cref{fig:model_structure}.
For the $i^{th}$
sample
\begin{small}
$X_i\in \mathbb{R}^{N\times 1\times T}$,
\end{small}
the initial $1\times 1$ convolution layer increases its hidden feature dimension to $d_{h1}$, outputting
\begin{small}
$(N, d_{h1}, T)$.
\end{small}
The encoder then encodes temporal and spatial information alternately, and generates a hidden representation of size
\begin{small}
$(N, d_{h2}, 1)$.
\end{small}
The encoder is followed by two linear layers to perform pooling on node embeddings and two MLP layers for classification. Cross entropy is used as the loss function:
\begin{small}
$L_{CE} = -\sum_i y_i \operatorname{log}\hat{y}_i$,
\end{small}
where $y_i\in \mathbb{R}^C$ is the one-hot vector of ground truth task labels and $\hat{y}_i\in \mathbb{R}^C$ is the model's predicted distribution.
We now explain the different components of the model.

\textbf{(I) Learning sample-level latent graph structures.}
Structural scans serve as our graph adjacency matrices. However, they remain fixed across temporal frames and across tasks. In contrast, FC (functional connectivities) are highly dynamic, resulting in different connection patterns across both time and tasks.
To better capture dynamic graph structures, we learn an adaptive adjacency matrix from each input graph signal. Unlike other works such as \cite{wu2019graph} that use a universal latent graph structure, our model does not assume that all samples share the same latent graph. Instead, our goal is to give each sample a unique latent structure that can reflect its own signaling pattern.
This implies that the latent adjacency matrix cannot be directly treated as a learnable parameter as a part of the model. To solve this, we minimize the assumption down to a shared projection $\Theta_{\text {adp}}$ that projects each input sequence into an embedding space and use this embedding to generate the latent graph structure. Projection $\Theta_{\text {adp}}$ can be learned in an end-to-end manner. The generated adaptive adjacency matrix for the $i^{th}$ sample can be written as follows ($\operatorname{Softmax}$ is applied column-wise):
\begin{small}
\begin{equation}
    A_{i\_\text{adp}}=\operatorname{Softmax}\left(\operatorname{ReLU}\left(\left(X_{i} \Theta_{\text {adp }}\right)\left(X_{i} \Theta_{\text {adp }}\right)^{\top}\right)\right), \Theta_{\text {adp }} \in \mathbb{R}^{T \times h_{\text {adp }}}
    \label{eq:adp}
\end{equation}
\end{small}

\textbf{(II) Gated TCN (Temporal Convolutional Network).}
To encode signal dynamics, we use the gating mechanism as in \cite{oord2016conditional} in our temporal layers:
\begin{small}
\begin{equation}
    H^{(l+1)} = \tanh\left(\operatorname{TCN}_{\text{emb}}(H^{(l)})\right) \odot \sigma\left(\operatorname{TCN}_{\text{gate}}(H^{(l)})\right),
\end{equation}
\end{small}
where $H^{(l)}\in\mathbb{R}^{N\times d\times t}$ is one sample's activation matrix of the $l^{th}$ layer, $\odot$ denotes the Hadamard product, and $\sigma$ is the Sigmoid function.
In contrast to TCNs that are generally used in sequence to sequence models that consist of dilated $\operatorname{Conv1d}$ and causal padding along the temporal dimension (\cite{oord2016wavenet}), we simply apply $\operatorname{Conv1d}$ with kernel = 2 and stride = 2 as our $\operatorname{TCN}_{\text{emb}}$ and $\operatorname{TCN}_{\text{gate}}$ to embed temporal information.
The reason is twofold: 
first, for a sequence to sequence model with a length-$T$ output, $y_\tau$ should only depend on $x_{t\leq\tau}$ to avoid information leakage and causal convolution can ensure this. In contrast, our model's task is classification, and the goal of our encoder along the temporal dimension is to embed signal information into the feature axis while reducing the temporal dimension to 1.
The receptive field of this single temporal point (with multiple feature channels) is meant to be the entire input sequence. Essentially, our TCN is the same as the last output node of a \textit{kernel-two causal TCN} whose dilation increases by two at each layer (\cref{fig:tcn}).
Second, from a practical perspective, directly using strided non-causal TCN works the same as taking the last node of dilated causal TCNs, as discussed above,
while simplifying the model structure and reducing training time to less than a quarter.

\textbf{(III) Graph Network layer.}
In our model, every set of $l$ temporal layers
(\cref{sssc:ab_study}
studies the best $l$ to choose) is followed by a spatial layer to encode signals with the graph structure. Building temporal and spatial layers alternately helps spatial modules to learn embeddings at different temporal scales,
and this generates better results than placing spatial layers after all the temporal ones.

To encode spatial information, \cite{kipf2017semisupervised} uses first-order approximation of spectral filters to form the layer-wise propagation rule of a GCN layer:
\begin{small}
$H^{(l+1)} = \operatorname{GCN}(H^{(l)}) = f(AH^{(l)}W^{(l)})$.
\end{small} It can be understood as spatially aggregating information among neighboring nodes to form new node embeddings. In the original setting without temporal signals, \begin{small}
$H^{(l)} \in \mathbb{R}^{N\times d}$
\end{small} is the activation matrix of $l^{th}$ layer, \begin{small}
$A\in \mathbb{R}^{N\times N}$
\end{small} denotes the normalized adjacency matrix with self-connections as discussed in \cref{sc:preliminaries}, \begin{small}
$W^{(l)}\in\mathbb{R}^{d\times d'}$
\end{small} is learnable model parameters, and $f$ is a nonlinear activation function of choice. Parameters $d$ and $d'$ are the number of feature channels.

We view a GCN layer as a local smoothing operation followed by an MLP, and simplify stacking K layers to \begin{small}$A^KH$\end{small} as in \cite{wu2019simplifying}.
In ReBraiD, every graph network layer aggregates information from each node's K-hop neighborhoods based on both brain structural connectivity and the latent adaptive adjacency matrix: thus, we have both \begin{small}$A_{i}{}^{K} H^{(l)}W_{K}$\end{small} and \begin{small}$A_{i\_\text{adp}}{}^{K} H^{(l)} W_{K\_\text{adp}}$\end{small} for input $H^{(l)}$. We also gather different levels (from $0$ to $K$) of neighbor information with concatenation.
In other words, one graph convolution layer here corresponds to a small module that is equivalent to K simple GCN layers with residual connections.
We can write our layer as:
\begin{small}
\begin{equation}
\label{eq:gnn}
    \begin{gathered}[b]
    H^{(l+1)}=\operatorname{GNN}^{(l)}\left(H^{(l)}\right)\\
    =\operatorname{MLP}\left[\operatorname{Concat}_{k=1}^{K}\left(H^{(l)}, \operatorname{ReLU}(A_{i}{}^{k} H^{(l)}), \operatorname{ReLU}(A_{i\_\text{adp}}{}^k H^{(l)})\right)\right] 
    \end{gathered}
\end{equation}
\end{small}
Note that in \cref{eq:gnn}, \begin{small}$A_i\in \mathbb{R}^{N\times N}$\end{small} and \begin{small}$H^{(l)}\in \mathbb{R}^{N\times d\times t}$\end{small}, and as a result their product \begin{small}$ \in \mathbb{R}^{N\times d \times t}$\end{small}.
Outputs of different $\operatorname{GNN}^{(l)}$ layers are parameterized and then skip connected with a summation. Since the temporal lengths of these outputs are different because of $\operatorname{TCN}$s, max-pooling is used before each summation to make the lengths identical.

\textbf{(IV) Multi-resolution inner cluster smoothing.} While $\operatorname{GNN}$ layers can effectively pass information between neighboring nodes, long-range relationships among brain regions that neither appear in SC nor learned by latent $A_{\text{adp}}$ can be better captured using soft assignments, similar to \textsc{DiffPool}\cite{ying2018hierarchical}.
To generate the soft assignment tensor $S^{(l)}$ that assigns $N$ nodes into $c$ clusters ($c$ chosen manually), we use \begin{small}
$\operatorname{GNN}_{pool}^{(l)}$
\end{small} that obeys the same propagation rule as in \cref{eq:gnn}, followed by $\operatorname{Softmax}$ along $c$. This assignment is applied to $Z^{(l)}$, the output of \begin{small}
$\operatorname{GNN}_{emb}^{(l)}$
\end{small} which carries out the spatial embedding for the $l^{th}$ layer input $H^{(l)}$, producing clustered representation $\tilde{H}^{(l)}$:
\begin{small}
\begin{equation}
    \begin{aligned}[c]
    &S^{(l)}=\operatorname{Softmax} \left(\operatorname{GNN}_{pool}^{(l)}\left(H^{(\ell)}\right), 1\right) \in \mathbb{R}^{N \times c \times t}\\
    &Z^{(l)}=\operatorname{GNN}_{emb}^{(l)}\left(H^{(l)}\right) \in \mathbb{R}^{N \times d \times t}\\
    &\tilde{H}^{(l)}= S^{(l) \top}Z^{(l)} \in \mathbb{R}^{c \times d \times t}\\
    \end{aligned}
\label{eq:diffpool}    
\end{equation}
\end{small}
The additional temporal dimension allows nodes to be assigned to heterogeneous clusters at different frames. 
We find that using coarsened \begin{small}$A_i^{(l+1)}=S^{(l)\top} A_i^{(l)} S^{(l)} \in \mathbb{R}^{c\times c}$\end{small} as the graph adjacency matrix leads to worse performance compared to using SC-generated $A_i$ and learned $A_{i\_\text{adp}}$
(comparison in \cref{ssc:model_components}).
In addition, if the number of nodes is changed, residual connections coming from the beginning of temporal-spatial blocks can not be used, impacting the overall performance. To continue using $A_i$ and $A_{i\_\text{adp}}$ as graph adjacency matrices and to allow residual connections, we reverse-assign $\tilde{H}^{(l)}$ with assignment tensor obtained from applying $\operatorname{Softmax}$ on $S^{(l) \top}$ along $N$, so that the number of nodes is kept unchanged:
\begin{small}
\begin{equation}
    \begin{aligned}[c]
    &\tilde{S}^{(l)}=\operatorname{Softmax} \left(S^{(l) \top}, 1\right) \in \mathbb{R}^{c \times N \times t}\\
    &H^{(\ell+1)}= \tilde{S}^{(l)^{\top}}\tilde{H}^{(l)} \in \mathbb{R}^{N \times d \times t}
    \end{aligned}
\label{eq:smoothing}
\end{equation}
\end{small}
In fact, \cref{eq:diffpool,eq:smoothing} perform signal smoothing on nodes within each soft-assigned cluster.
With the bottleneck $c < N$, the model is forced to pick up latent community structures.
This inner cluster smoothing is carried out at multiple spatial resolutions: as the spatial receptive field increases with more graph layers, we decrease cluster number $c$ for the assignment operation.
As these $\operatorname{GNN}$ layers alternate with $\operatorname{TCN}$ layers, the inner cluster smoothing also learns the community information across multiple temporal scales.

\subsection{Attribution with IG (Integrated Gradients)} 
As one approach to model interpretability, \textit{attribution} assigns credits to each part of the input, assessing how important they are to the final predictions.
\cite{wiltschko2020evaluating} gives an extensive comparison between different graph attribution approaches, in which IG \cite{sundararajan2017axiomatic} is top-performing and can be applied to trained models without any alterations of the model structure. IG also has other desirable properties, such as implementation invariance that other gradient methods lack.
It is also more rigorous and accurate than obtaining explanations from attention weights or pooling matrices that span multiple feature channels. Intuitively, IG calculates how real inputs contribute differently compared to a selected baseline; it does so by aggregating model gradients at linearly interpolated inputs between the real and baseline inputs. 

In order to apply IG, we calculate attributions at each point of both input \begin{small}$A\in\mathbb{R}^{N\times N}$\end{small} and \begin{small}$X\in\mathbb{R}^{N\times T}$\end{small} for each sample:
\begin{small}
\begin{equation}
    \begin{aligned}
    & \operatorname{\textsc{Attr}}_{\mathcal{G}_{vw}} =\left(\mathcal{G}_{vw}-\mathcal{G}_{vw}^{\prime}\right) \times 
    \sum_{m=1}^{M} \frac{\partial F\left(\mathcal{G}_{\text{Intrpl}}\right)}{\partial \mathcal{G}_{\text{Intrpl}_{vw}}} 
    \times\frac{1}{M},\\
    & \mathcal{G} = (A, X), \quad \mathcal{G}_{\text{Intrpl}} = \mathcal{G}^{\prime}+
    \frac{m}{M}\times\left(\mathcal{G}-\mathcal{G}^{\prime}\right)
    \end{aligned}
\label{eq:ig}    
\end{equation}
\end{small}
\begin{small}
$F(\mathcal{G})$
\end{small} here represents our signal classification model, $M$ is the step number when making Riemann approximation of the path integral, and $\mathcal{G}^{\prime}$ is the baselines of $\mathcal{G}$ (see \cref{ssc:interpretations}
for more details). Note that \cref{eq:ig} calculates the attribution of one edge or one node on one sample. 
The process is repeated for every input point, so attributions $\operatorname{\textsc{Attr}}_A, \operatorname{\textsc{Attr}}_X$ have identical dimensions as inputs $A, X$.
To obtain the brain region importance of a task, we aggregate attributions across multiple samples of that task.

\section{Experiments}
We use fMRI signals from the CRASH dataset \cite{lauharatanahirun2020flexibility} for our experiments. The model classifies input fMRI into six tasks: resting state, VWM (visual working memory task), DYN (dynamic attention task), MOD (math task), DOT (dot-probe task), and PVT (psychomotor vigilance task). We preprocess 4D voxel-level fMRI images into graph signals $\mathcal{G} = (A, X)$ by averaging voxel activities into regional signals with the 200-ROI cortical parcellation (voxel to region mapping) specified by \cite{schaefer2018local}. We also standardize signals for each region and discard scan sessions with obvious abnormal spikes that may be caused by head movement, etc. DWI scans are mapped into the same MNI152 coordinate and processed into adjacency matrices with the same parcellation as fMRI. Our processed data contains 1940 scan sessions from 56 subjects. Session length varies from 265 frames
to 828 frames (see \cref{tab:data_detail} for details).
TR (Repetition Time) is 0.91s.

The 1940 scan sessions from CRASH are separated into training, validation, and test sets with a ratio of 0.7-0.15-0.15 (subject-wise split does not lead to any noticeable difference). Each split receives a proportional number of samples for each class. Hyperparameters including dropout rate, learning rate, and weight decay are selected using grid search based on validation loss. All results reported in this section are obtained from the test set. For each scan session, we use a stride-10 sliding window to generate input sequences (in the following experiments $T \in \{8, 16, 32, 64, 128, 256\}$) and feed them to the model. To encode temporal and spatial information alternately, we find stacking two $\operatorname{TCN}$ layers per one $\operatorname{GNN}$ layer leads to better performance most times (see \cref{sssc:ab_study} (I)). 
We tested $h_\text{adp} = 2, 5, 10$ in \cref{eq:adp} for our experiments, and 5 appears to be the best; so we use this value for all the following experiments. $K = 1,2,3$ in \cref{eq:gnn}  were tested on a few settings, and K = 2, 3 have a similar performance, both outperforming K = 1. Since smaller values of K have smaller computation needs, we use $K = 2$ for all experiment settings, meaning each $\operatorname{GNN}$ layer aggregates information from 2-hop neighbors based on the provided adjacency matrices. We evaluate our model with weighted F1 as the metric in order to account for the imbalance in the number of samples in each task.
Our models are written in PyTorch, trained with Google Colab GPU runtimes, and 30 epochs are run for each experiment setting.
Code is publicly available \footnote{\href{https://github.com/sklin93/ReBraiD}{https://github.com/sklin93/ReBraiD}}.

\begin{table}
\caption{fMRI scan details for six tasks.}
\vspace{-0.1cm}
\label{tab:data_detail}
\resizebox{0.48\textwidth}{!}{%
\begin{tabular}{c|cccccc|c}
\toprule
Tasks               & Rest & VWM & DYN & DOT & MOD & PVT & (Total) \\\hline
Valid sessions & 209     & 514 & 767 & 155 & 138 & 157 & 1940    \\
Frames / Session       & 321     & 300 & 265 & 798 & 828 & 680 & ---
\\\bottomrule
\end{tabular}}
\end{table}

\subsection{Model components}
\label{ssc:model_components}
\begin{figure}[t]
    \centering
    \captionsetup{font=small}
    \centering
    \vspace{-0.2cm}
    \includegraphics[ width=0.47\textwidth]{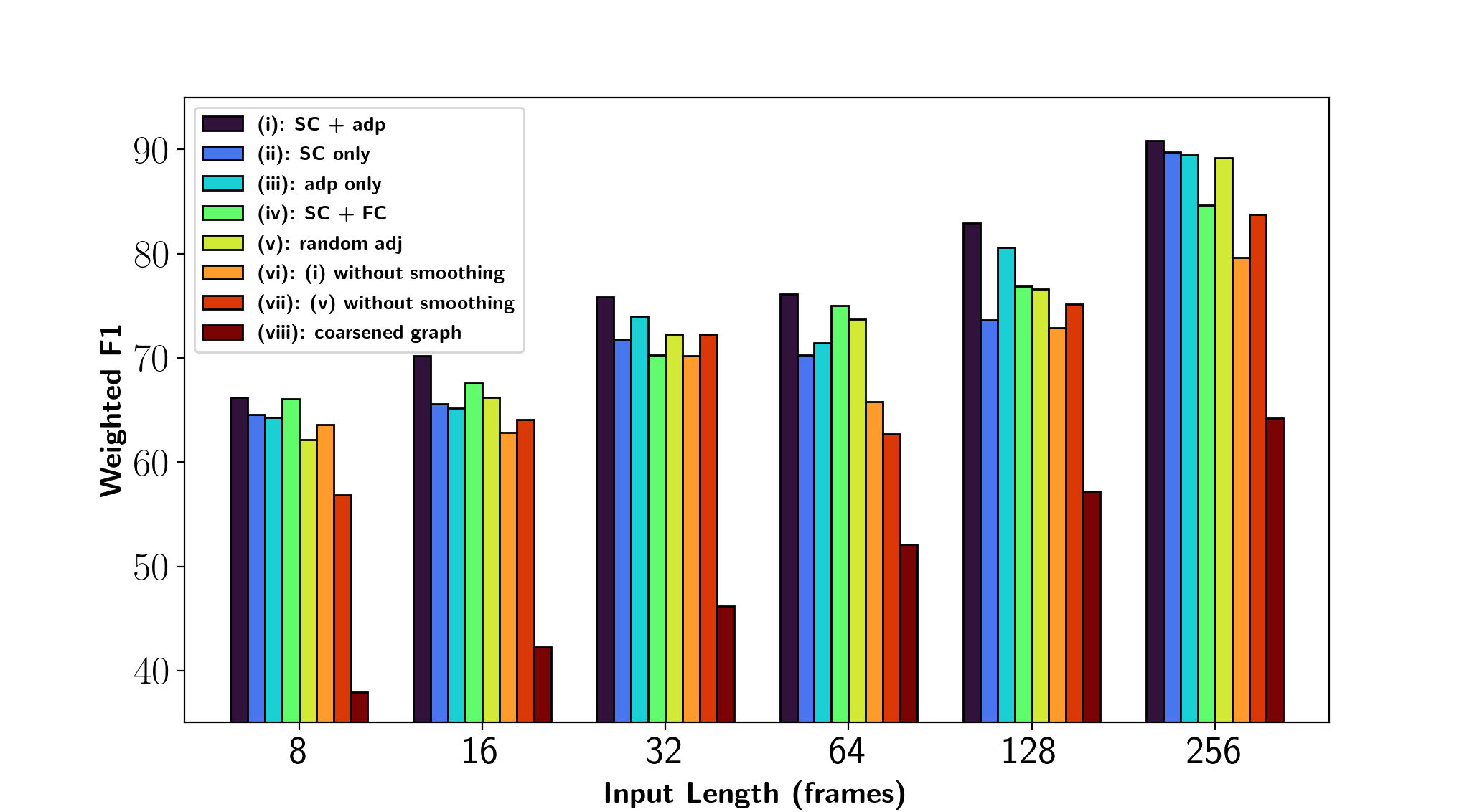}
    
    \vspace{-0.1cm}
    \caption{
    Ablation studies on different input length (please see \cref{tab:ab_study} in appendix for numerical values of weighted F1 under each setting).
    }
    \label{fig:graph_supports}
\end{figure}

\textbf{Ablation studies on graph adjacency matrices.}
For each input sample $\mathcal{G}_i$, we test different options to provide graph adjacency matrices to the $\operatorname{GNN}$ layer. They include (i) our proposed method: using both adaptive adjacency matrix $A_{i\_\text{adp}}$ and SC-induced $A_i$, (ii) only using $A_{i}$, (iii) only using $A_{i\_\text{adp}}$, (iv) replacing $A_{i\_\text{adp}}$ in setting i with $A_{i\_\text{FC}}$ derived from functional connectivity, and (v) only using random graph adjacency matrices with the same level of sparsity as real $A$'s.
The results under different settings are reported in \cref{fig:graph_supports} (and \cref{tab:ab_study} in appendix for numerical values).

From the results of setting (ii) plotted in \cref{fig:graph_supports}, we see that removing the adaptive adjacency matrix impacts the performance differently at different input lengths: the gap peaks for signals of length 64--128, and becomes smaller for either shorter or longer sequences. This could suggest the existence of more distinct latent states of brain signals of this length that structural connectivities cannot capture.
On the other hand, removing SC (setting (iii)) seems to have a more constant impact on the model performance, with shorter inputs more likely to see a slightly larger drop. In general, only using $A_{\text{adp}}$ leads to a smaller performance drop than only using SC,
indicating the effectiveness of $A_{\text{adp}}$ in capturing useful latent graph structures. More detailed studies below show that $A_{\text{adp}}$ learns distinct representations not captured by $A$.

As mentioned in \cref{sc:method}, our motivation behind creating sample-level adaptive adjacency matrices is FC's highly dynamic nature. Therefore, for setting (iv), we test directly using adjacency matrices $A_{i\_\text{FC}}$ obtained from FC instead of the learned $A_{i\_\text{adp}}$. In particular,
\begin{small}
$A_{i\_\text{FC}} =  \tilde{D}_{\text{FC}_i}^{-\frac{1}{2}}\tilde{\text{FC}_i}\tilde{D}_{\text{FC}_i}^{-\frac{1}{2}}\in \mathbb{R}^{200\times 200}$,
\end{small} where
\begin{small}
$(\text{FC}_i)_{vw} = \operatorname{corr}((X_i)_v, (X_i)_w)$,
$\tilde{\text{FC}_i}=\text{FC}_i+I_N$
\end{small} and
\begin{small}
$\tilde{D}_{\text{FC}_i}=\sum_w (\tilde{\text{FC}_i})_{vw}$.
\end{small}
Fig. \ref{fig:graph_supports} shows $A_{i\_\text{FC}}$ constantly underperforms $A_{i\_\text{adp}}$, except for being really close for length-8 inputs. Larger performance gaps are observed for longer inputs, where \begin{small}$\operatorname{Corr}((X_i)_v, (X_i)_w)$ \end{small} struggles to capture the changing dynamics in the inputs. This demonstrates that our input-based latent $A_{i\_\text{adp}}$ has better representation power than input-based FC. We also notice batch correlation coefficients calculation for $A_{i\_\text{FC}}$ results in a slower training speed than computing $A_{i\_\text{adp}}$.

An interesting result comes from setting (v), where we use randomly generated Erdős-Rényi graphs with the edge creation probability the same as averaged edge existence probability of $A$'s. Its performance is similar to or even better than settings (ii) and (iii).
We examine this further in \cref{ssc:interpretations}.


\textbf{Latent adaptive adjacency matrix $A_{\text{adp}}$.} The above results demonstrate latent $A_{\text{adp}}$ can complement the task- and temporal-fixed $A$. We now show that the learned $A_{i\_\text{adp}}$ is sparse for each sample, has evident task-based patterns, and provides new information beyond $A_i$.
The sparsity of $A_{i\_\text{adp}}$ can be seen from  \cref{fig:adp_ind_task3} in appendix: each input only gets a few important columns (information-providing nodes in $\operatorname{GNN}$). These columns vary from one sample to another, indicating $A_{\text{adp}}$'s ability to adapt to changing inputs within the same task. However, when we look into inputs generated by consecutive sliding windows (not shuffled) from the same scan session as in \cref{fig:adp_ind_task3_same_ses}, we can see the latent structures change smoothly.
In addition, when we aggregate samples inside each task, noticeable task-based patterns emerge (\cref{fig:adp_task_avg_roi}). These patterns are different from $\operatorname{Attr}_A$ in \cref{fig:attr_A_with_rand_colsum}, suggesting that $A_{\text{adp}}$ embeds dynamics not captured by $A$.

Quantitatively, $A_{i\_\text{adp}}$ entry values range between (0, 1) because of the $\operatorname{Softmax}$, and only around 2\% of entries in $A_{i\_\text{adp}}$ have values larger than 0.05. As a reference, the largest entry value is larger than 0.99. A similar sparsity pattern is found when using synthetic data on the same model, indicating that the sparsity is more due to the model than the underlying biology. Given how $A_{i\_\text{adp}}$ is used in GNN layers, each column of it represents a signal-originating node during message passing. We hypothesize that the model learns the most effective \textit{hubs} that pass information to their neighbors. A related idea is information bottleneck \cite{tishby2015deep}: deep learning essentially compresses the inputs as much as possible while retaining the mutual information between inputs and outputs. In a sense, $A_{i\_\text{adp}}$ represents the compressed hubs for a given input signal.
We also note that this sparsity emerges even without any additional constraints. In fact, adding $L_1$ constraints on $A_{\text{adp}}$ does not change the model performance or the $A_{i\_\text{adp}}$ sparsity level. We hypothesize that the naturally trained $A_{i\_\text{adp}}$ is sparse enough, and further sparsification is unnecessary.

We visualize the projected inputs $X_i \Theta_{\text {adp }}$ in \cref{fig:nodevec}, which clearly shows the task, node and subject heterogeneities. Different tasks have varied representations in the latent space for the same node, but DOT, MOD, PVT has similar embedding patterns across individuals and most nodes. Indeed, when looking at the confusion matrix across models (\cref{fig:cm} in appendix), the misclassifications mostly cluster between these three tasks, indicating their natural similarity. We want to note here that adding a learnable bias to $X \Theta_{\text {adp }}$ does not separate the task embeddings further, nor does it improve overall performance.
Subjects also exhibit heterogeneity: the same pair of nodes during the same task can have different embedding distances, thus graph edge weights, for each individual.

\textbf{Multi-resolution inner cluster smoothing.}
To verify the capability of inner cluster smoothing operation in capturing latent graph dynamics, we test the following settings: (vi) using our proposed model and inputs, except removing paralleled $\operatorname{GNN}_{pool}$ and inner cluster smoothing module; (vii) previous setting (v) but remove $\operatorname{GNN}_{pool}$ and inner cluster smoothing module; (viii) keep $\operatorname{GNN}_{pool}$, but using coarsened graph instead of smoothing (essentially performing \textsc{DiffPool} with an added temporal dimension). In this last setting, we hierarchically pool and reduce the graph to a single node, and we keep the total number of $\operatorname{GNN}$ layers the same as our other settings.
Values of soft-assigned cluster number $c$ are chosen to be halved per smoothing module (e.g., $N/2, N/4, \cdots$) for our experiments. Different choices of $c$ affect the model convergence rate but only have a minor impact on the final performance (see \cref{sssc:ab_study} (II)).
Results are reported in \cref{fig:graph_supports} (and \cref{tab:ab_study} in appendix). Apart from these three settings, we also test adding pooling regularization terms (described in \cref{ssc:pool_reg}) into the loss function, but they do not lead to much of a difference.

The above results demonstrate that both setting (vi) and (vii) outperforms (viii) by a large margin, indicating the importance of keeping the original node number when representing brain signals. In addition, all three settings underperform our proposed method. They are also mostly worse than changing graph adjacency matrices as in settings (ii)--(v): this shows the inner cluster smoothing module has a more significant impact in learning latent graph dynamics.
We also find using adaptive adjacency matrices and inner cluster smoothing can stabilize training, making the model less prone to over-fitting and achieving close-to-best performance over a larger range of hyperparameters (see \cref{fig:c_effect}).

\subsection{Model Comparisons}
\label{ssc:model_comparison}
\begin{table}[t]
    \centering
    \caption{Model comparisons with length-256 inputs.}
    \label{tab:comparison}
    \vspace{-0.1cm}
    \resizebox{0.47\textwidth}{!}{%
    \begin{tabular}{c|c|c}
        \toprule
        Model & Weighted F1 & \multicolumn{1}{c}{\begin{tabular}[c]{@{}c@{}}Training time\\ (s / epoch)\end{tabular}} \\
        \hline
        GCN \cite{kipf2017semisupervised} & 42.84 & 713\\
        GAT V2 \cite{brody2021attentive} & 50.36 & 1142 \\
        GConvGRU \cite{seo2018structured} & 56.05 & 9886 \\
        GraphSAGE \cite{hamilton2017inductive} & 61.87 & 1048 \\
        Graph Transformer \cite{shi2021masked} & 66.11 & 1890 \\\hline
        MVTS Transformer \cite{zerveas2021transformer} & 88.16 & \textbf{39} \\\hline
        \textbf{ReBraiD} (proposed: TCN + GNN) & \textbf{90.85} & 298 \\
        ReBraiD (TCN only) & 71.98 & 119 \\
        ReBraiD (TCN + CNN) & 75.79 & 124
        \\\bottomrule
    \end{tabular}}
\end{table}
 
Since we adopt a network view to studying the brain, where brain regions are treated as graph nodes, we source our baselines from graph models. To do so, we examined all models in PyTorch Geometric (PyG) \footnote{\href{https://pytorch-geometric.readthedocs.io/}{https://pytorch-geometric.readthedocs.io/}} and its temporal extension (PyG-T) \footnote{\href{https://pytorch-geometric-temporal.readthedocs.io/}{https://pytorch-geometric-temporal.readthedocs.io/}} as they contain the most up-to-date and well-organized open-source graph neural network model implementations. In particular, we compare our model with the vanilla GCN from \cite{kipf2017semisupervised},
Chebyshev Graph Convolutional Gated Recurrent Unit (GConvGRU) from \cite{seo2018structured}, GraphSAGE from \cite{hamilton2017inductive}, GAT V2 from \cite{brody2021attentive}
and Graph Transformer as in \cite{shi2021masked}.
Baseline models are constructed similar to ours: each has four graph encoding layers taking in both signals and adjacency matrices, followed by two linear layers along the node axis and two linear layers for the final classification.
We train baseline models with the same input, loss, optimizer, and epoch settings (all models are well-converged). Grid search is used to optimize the rest of the hyperparameters. We compare weighted F1 and training time per epoch in \cref{tab:comparison}; we also plot our model and Graph Transformer's confusion matrices in \cref{fig:cm}.

Our model shows significant performance gains and requires less training time than graph baselines. We believe the most critical reason is that the models in PyG treat temporal signals as feature vectors instead of placing them into a separate temporal dimension. Without sequence modeling on the temporal dimension, even the state-of-the-art graph attention models (GAT-v2 and graph Transformer) cannot perform well. In addition, almost all models in PyG-T assume one common graph for the inputs (application scenarios are traffic network forecasting, link predictions, etc.), whereas we need to feed different SC for every sample. Out of them, we were able to choose one model (GConvGRU) that supports different adjacency matrices, but it didn't give a satisfactory result. Our proposed ideas of sample-level adaptive adjacency matrix learning and multi-resolution inner cluster smoothing help capture latent brain dynamics and improve the performance. The higher model performance here reflects a better encoding ability of brain signals, which can benefit different downstream tasks such as disease and trait prediction.

In addition to graph baselines, we also tested the state-of-the-art model for multivariate time series classification (MVTS Transformer \cite{zerveas2021transformer}), which has comparable performance to ours. This stresses the critical role of temporal modeling when dealing with dynamic signals, so we tested our model without GNN layers. We experiment both removing GNN layers altogether and replacing them with $1\times 1$ CNN layers: both outperform graph models that focus on the spatial modeling aspect. 
Although these results demonstrate that temporal modeling is crucial, adding graph modeling that includes signals' spatial relationships as proposed can further improve the performance. Since the MVTS Transformer model has projections to generate queries, keys, and values from the input sequence, it can also implicitly learn spatial relationships between variables (\textit{nodes}). On the other hand, explicitly adding graph components allows the model to utilize prior structures (e.g., SC). The attribution of graph models can also provide better interpretability of brain networks, such as identifying critical region connections.

\subsection{Interpretation with IG}
\label{ssc:interpretations}
This section studies the contributions of different brain ROIs and subnetworks defined by their functionalities. For the subnetwork definition, we choose to use the 17 networks specified in \cite{thomas2011organization}, which has a mapping from our previous 200-ROI parcellation\footnote{\href{https://github.com/ThomasYeoLab/CBIG/blob/master/stable_projects/brain_parcellation}{https://github.com/ThomasYeoLab/CBIG/blob/master/stable\_projects/brain\_parcellation}}.
To select baseline inputs, we follow the general principle for attribution methods: when the model takes in a baseline input, it should produce a near-zero prediction, and $\operatorname{Softmax}(\text{outputs})$ should give each class about the same probability in a classification model. 
All-zero baselines $A^{\prime}$ and $X^{\prime}$ can roughly achieve this for our model, so we choose them as our baseline inputs.
Step number $M$ is set to 30.
The IG computation is done on 900 inputs for each task to get an overall distribution.

The extracted high-attribution regions and connections should be reproducible across different initializations to be used for downstream tasks. Since the overall problem is non-convex, we empirically test and confirm the attribution reproducibility with two randomly initialized models before proceeding to the following analyses. In addition, \cite{wiltschko2020evaluating} demonstrates IG's consistency (reproducibility among a range of hyperparameters) and faithfulness (more accurate attribution can be obtained with better performing models). Since our model has higher performance with longer inputs, we compute IG attributions of a model trained on length-256 input signals in this section.

\begin{figure}[t]
    \centering
    \captionsetup{font=small}
    \begin{subfigure}[b]{0.14\textwidth}
    \includegraphics[width=\textwidth]{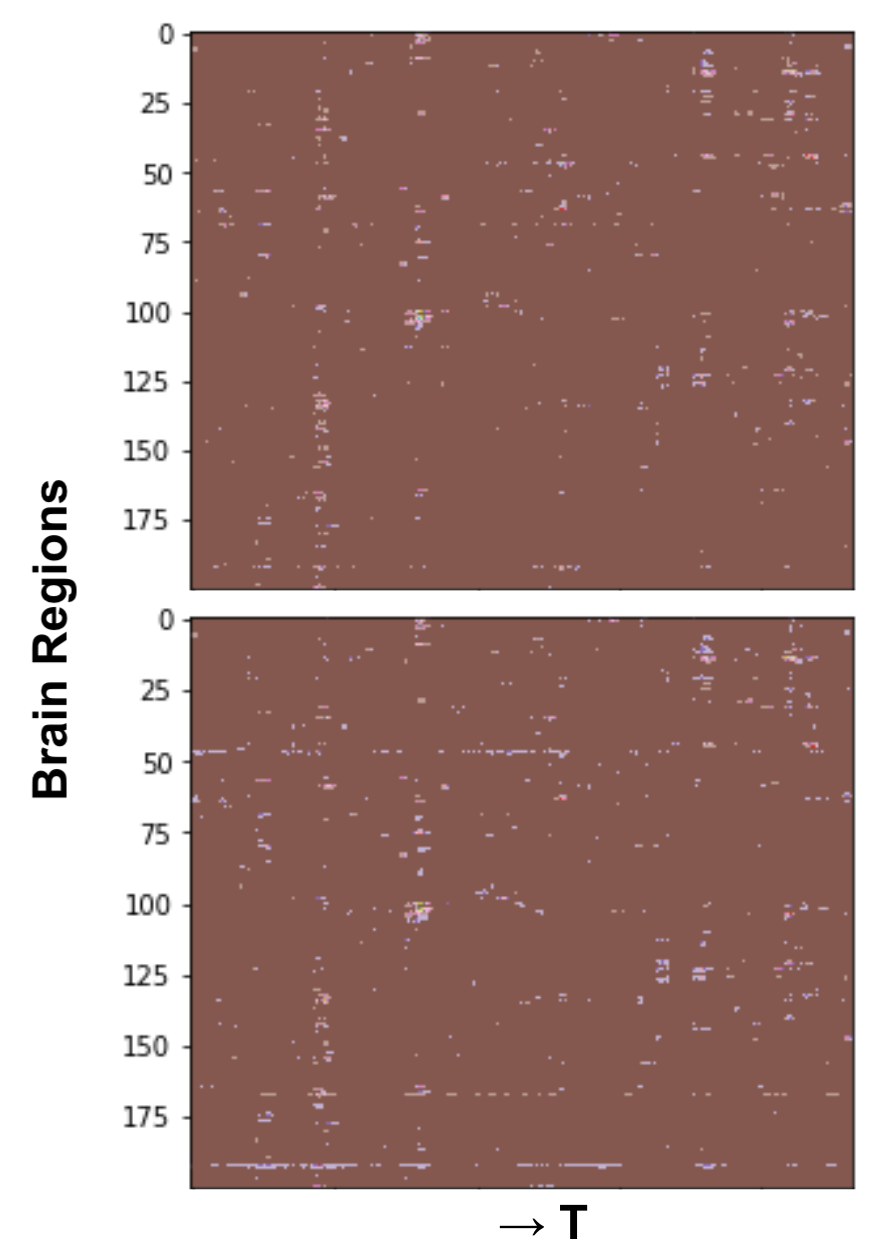}
    \caption{}
    \label{fig:ig_sanity_check}
    \end{subfigure}    
    \begin{subfigure}[b]{0.32\textwidth}
    \centering
    \includegraphics[width=\textwidth]{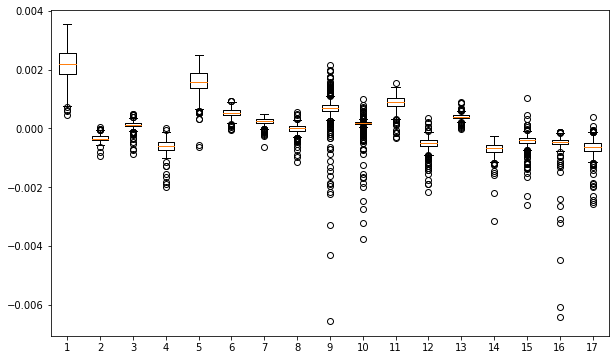}
    \caption{}
    \label{fig:ig_task}
    \end{subfigure}
    \caption{
    (a) Temporal importance sanity check of IG results on two pieces of inputs with a large overlap period. Attribution maps are offset aligned. 
    (b) $\operatorname{\textsc{Attr}}_{X}$ distributions across 17 brain subnetworks (defined as in \cite{thomas2011organization}) for VWM.
    }
    \label{fig:ig}
    \vspace{-0.5cm}
\end{figure}

\textbf{Temporal importance.}
On the single input level, we can attribute which parts of the inputs in $\mathcal{G}_i$ are more critical in predicting the target class by looking into $(\operatorname{\textsc{Attr}}_{X})_i$. This attribution map not only shows which brain regions contribute more but also reveals the important signal frames. One critical drawback of fMRI imaging is its low temporal resolution, but if we know which part is more important, we can turn to more temporally fine-grained signals such as EEG to see if there are any special activities during that time.
To confirm that the attributions we get are valid and consistent, we perform a sanity check of IG results on two overlapped inputs with an offset $\tau$: the first input is obtained from window $[t_0, t_0+T]$ and the second is obtained from window $[t_0+\tau, t_0+\tau+T]$. Offset aligned results are shown in \cref{fig:ig_sanity_check}, in which the attributions agree with each other quite well.

\begin{figure*}[t]
    \centering
    \captionsetup{font=small}
    \includegraphics[trim={10pt 0 6pt 0}, clip, width=0.9\textwidth]{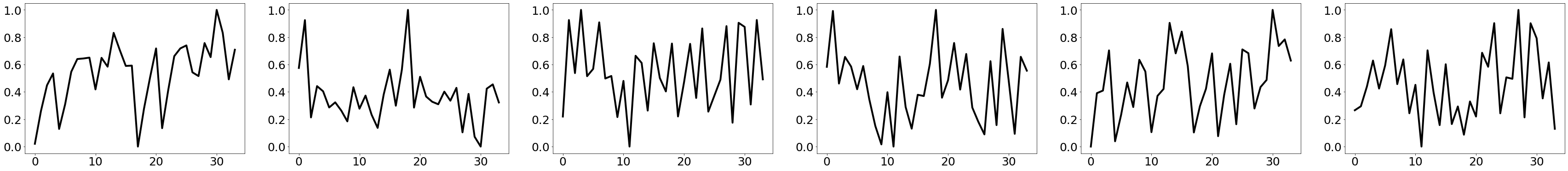}\\
    \includegraphics[trim={10pt 0 6pt 0}, clip, width=0.9\textwidth]{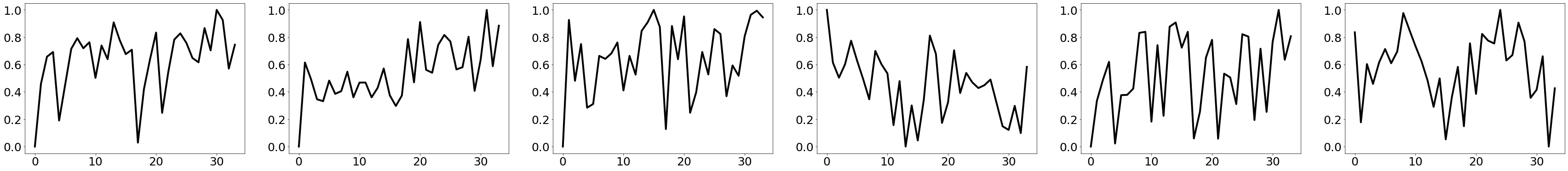}
    \caption{Column averages of task-averaged $\operatorname{\textsc{Attr}}_{A}$ (mapped into 34 subnetworks defined by the 17-network parcellation with left, right hemispheres). Top row is obtained from real SC induced $A$ and bottom rows is obtained from random SC induced $A_{\operatorname{rand}}$. Attributions are normalized to $[0, 1]$. Tasks are: Rest, VWM, DYN, DOT, MOD, PVT from left to right.}
    \label{fig:attr_A_with_rand_colsum}
\end{figure*}

\begin{figure*}[t]
    \centering
    \captionsetup{font=small}
    \includegraphics[width=0.15\textwidth]{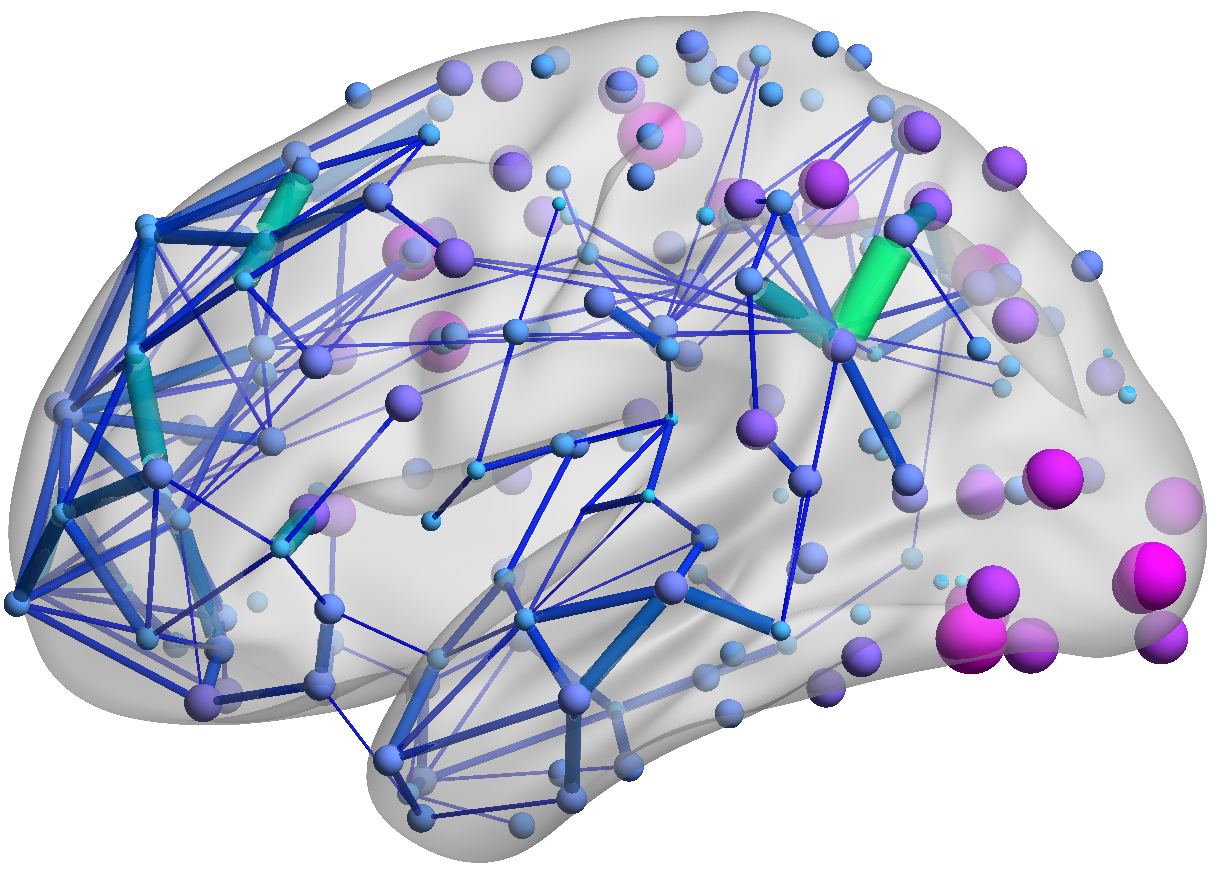}
    \includegraphics[width=0.15\textwidth]{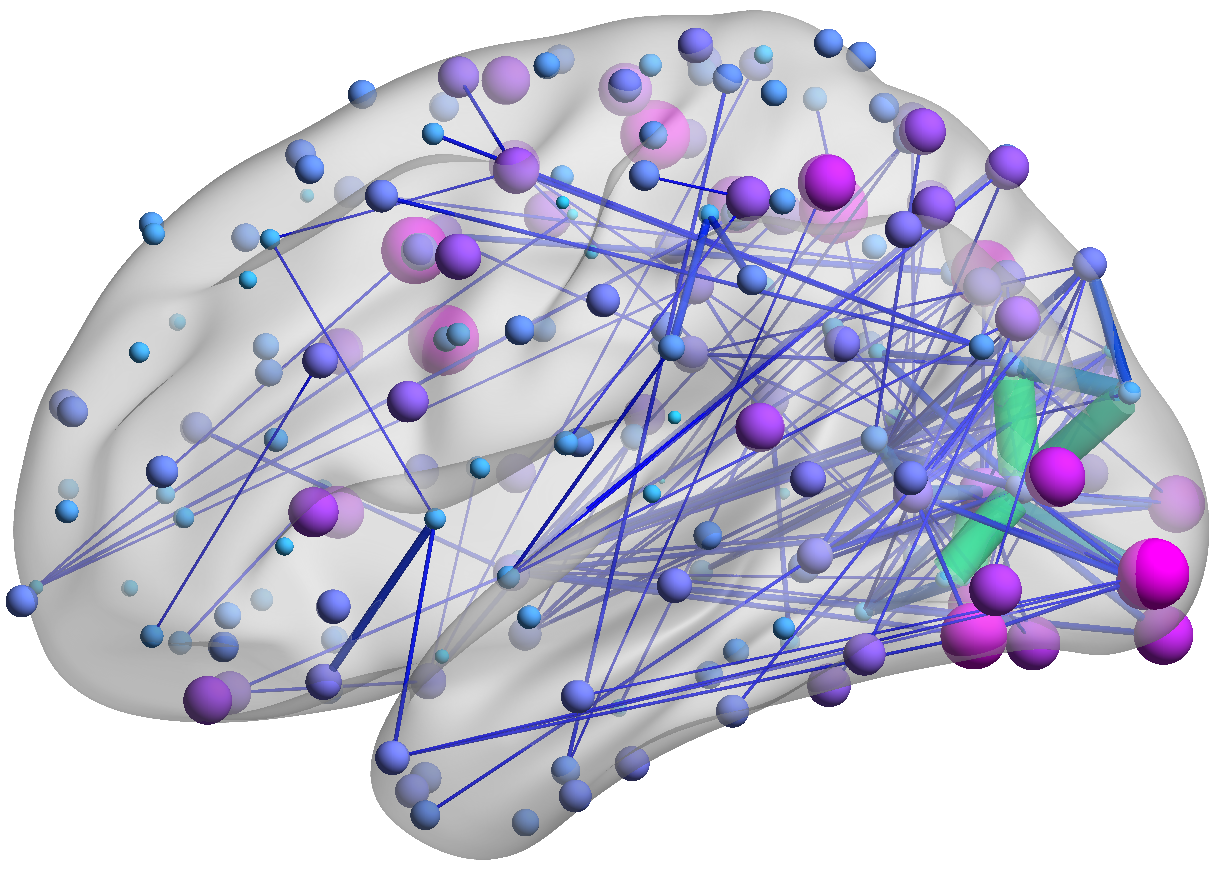}
    \includegraphics[width=0.15\textwidth]{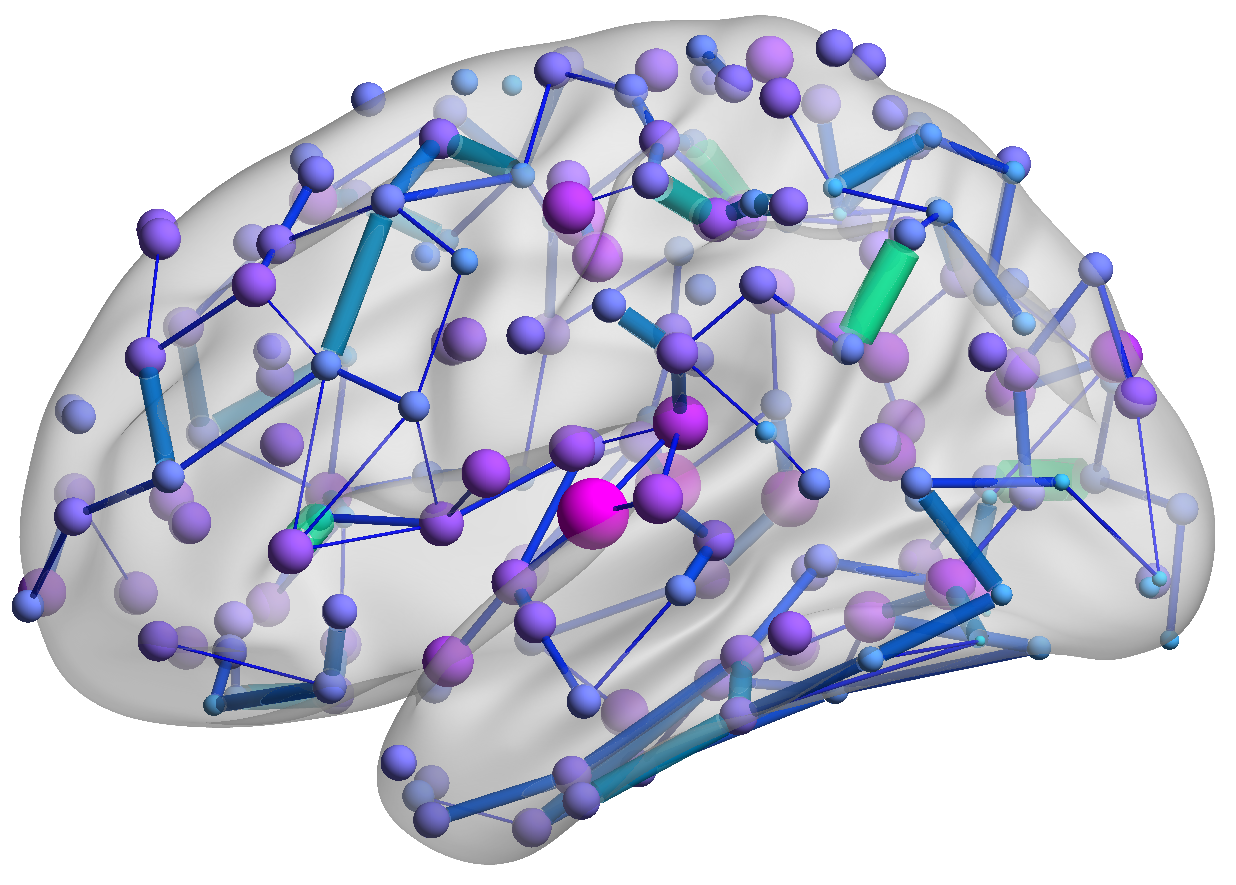}
    \includegraphics[width=0.15\textwidth]{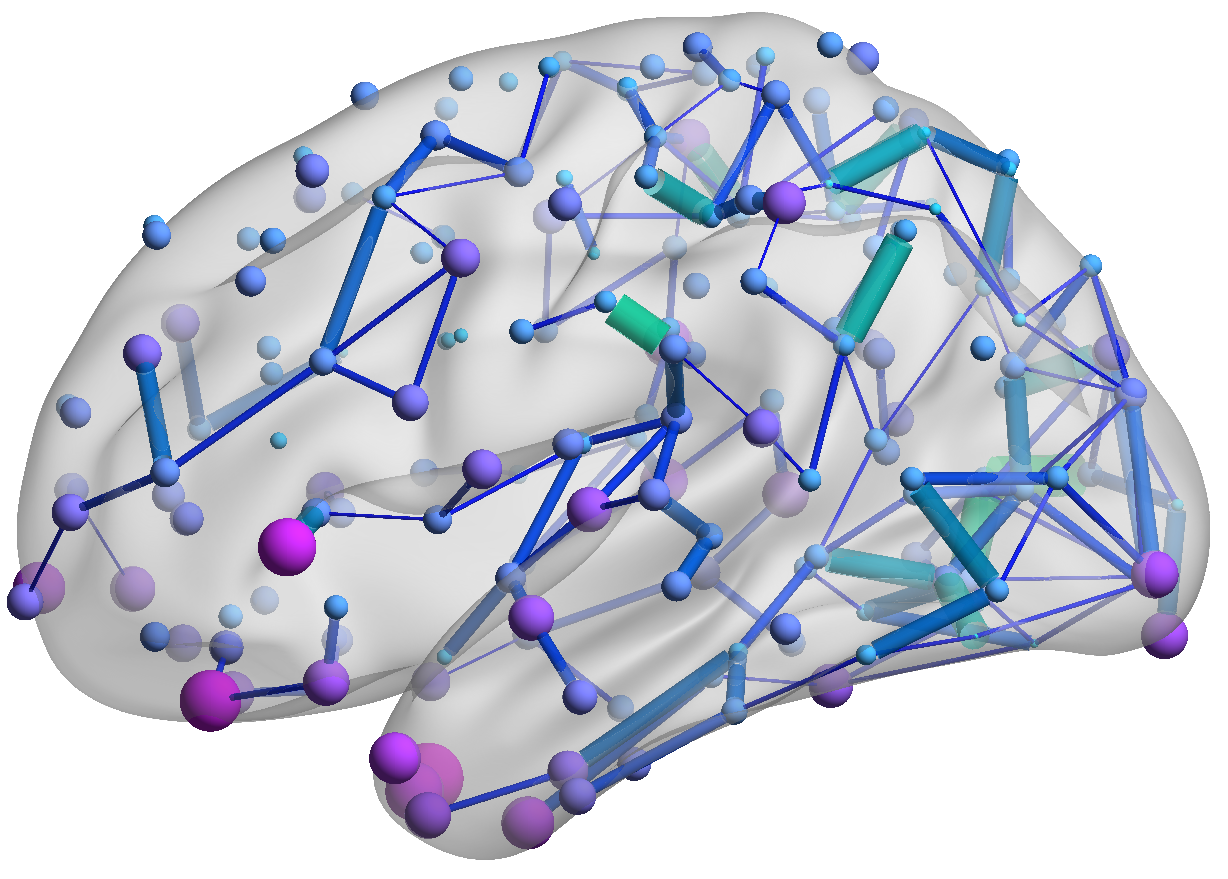}
    \includegraphics[width=0.15\textwidth]{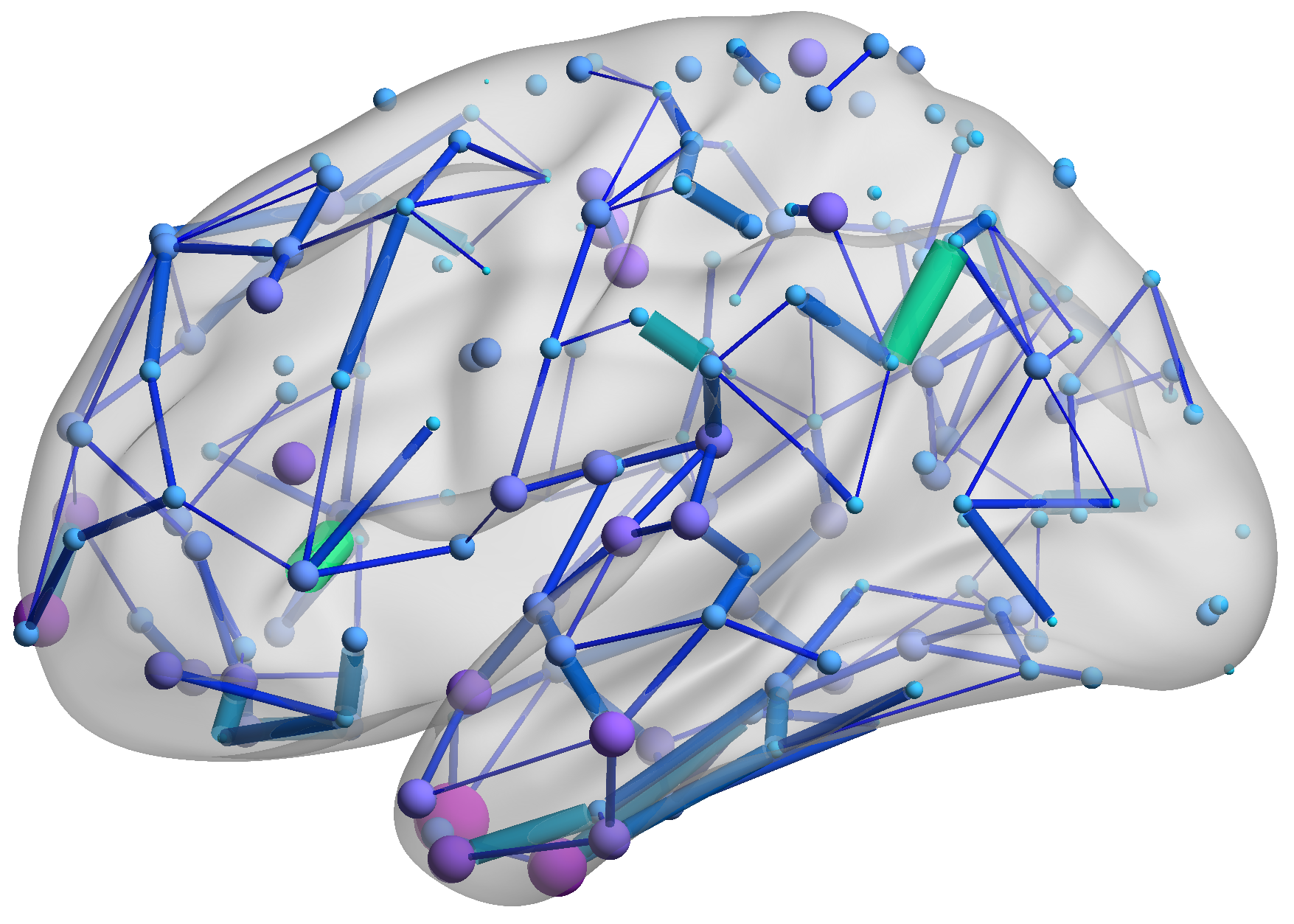}
    \includegraphics[width=0.15\textwidth]{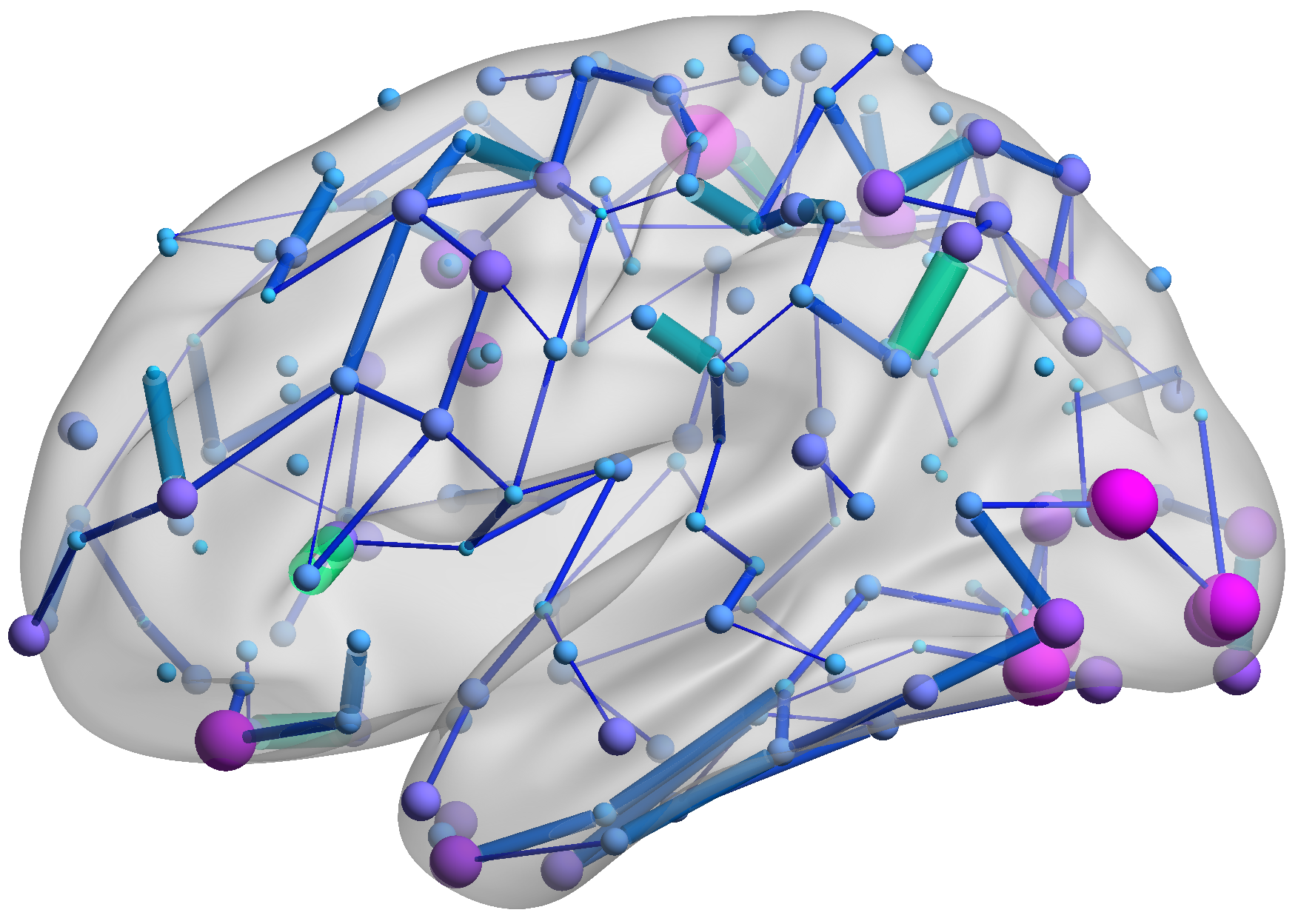}
    \includegraphics[width=0.015\textwidth]{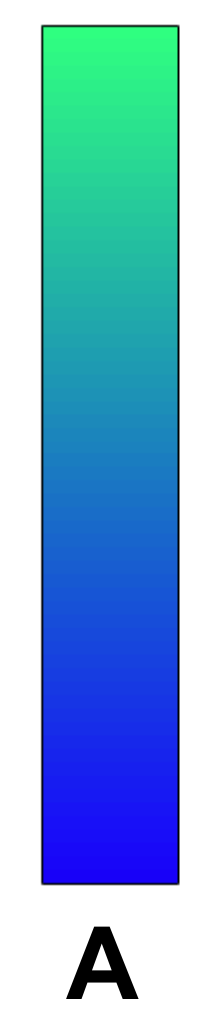}
    \\
    \includegraphics[width=0.115\textwidth]{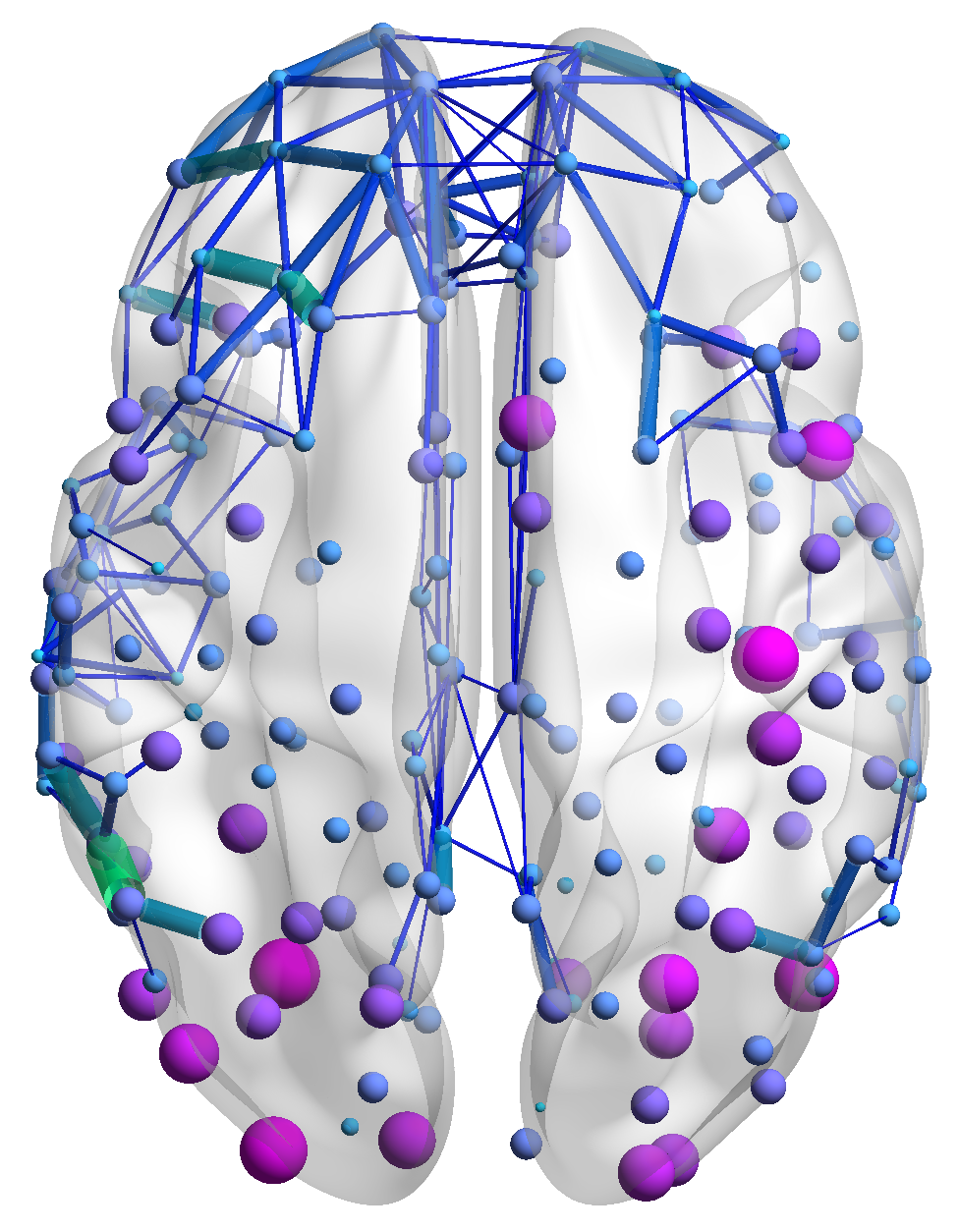}\hspace{+13pt}
    \includegraphics[width=0.115\textwidth]{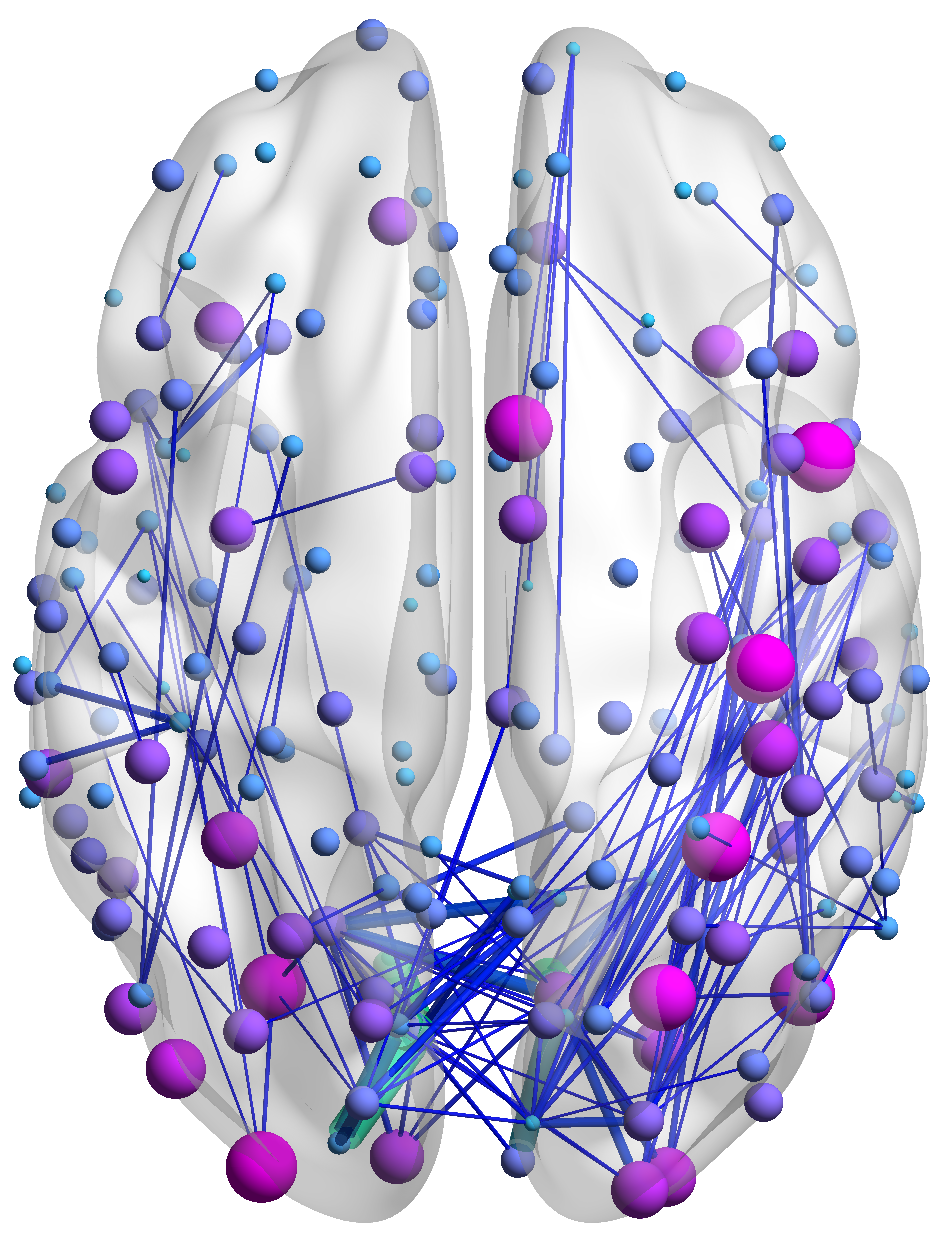}\hspace{+13pt}
    \includegraphics[width=0.115\textwidth]{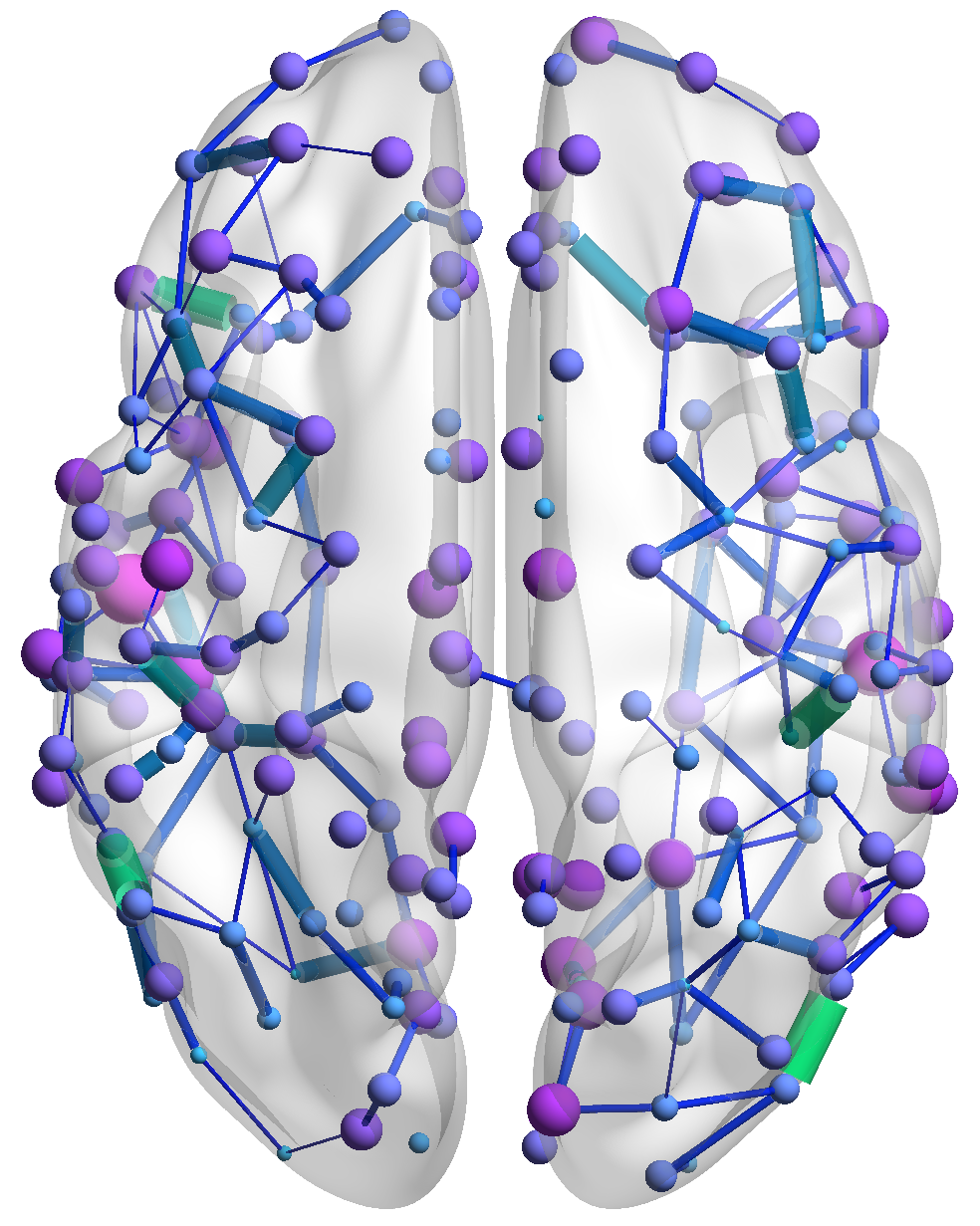}\hspace{+13pt}
    \includegraphics[width=0.115\textwidth]{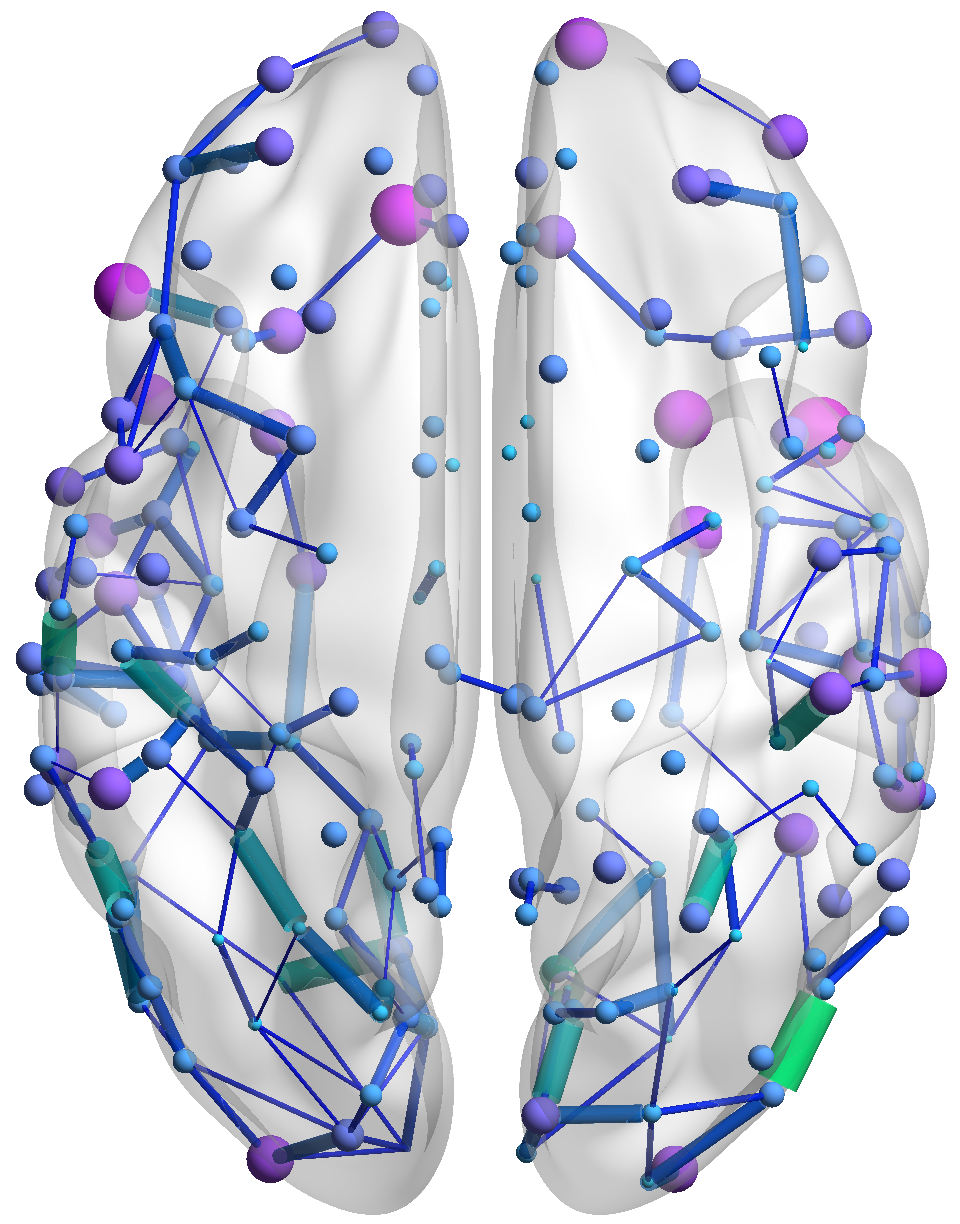}\hspace{+13pt}
    \includegraphics[width=0.115\textwidth]{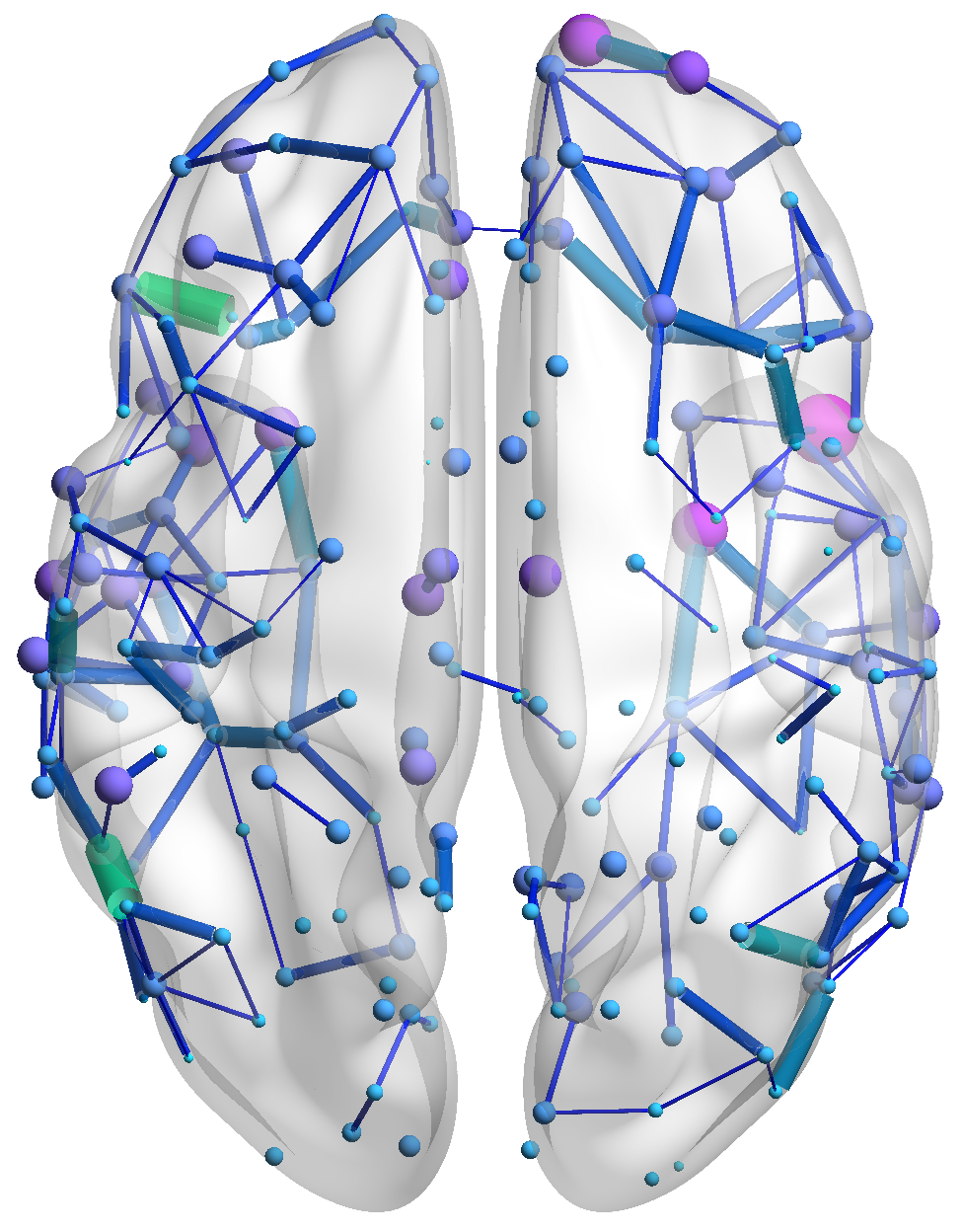}\hspace{+13pt}
    \includegraphics[width=0.115\textwidth]{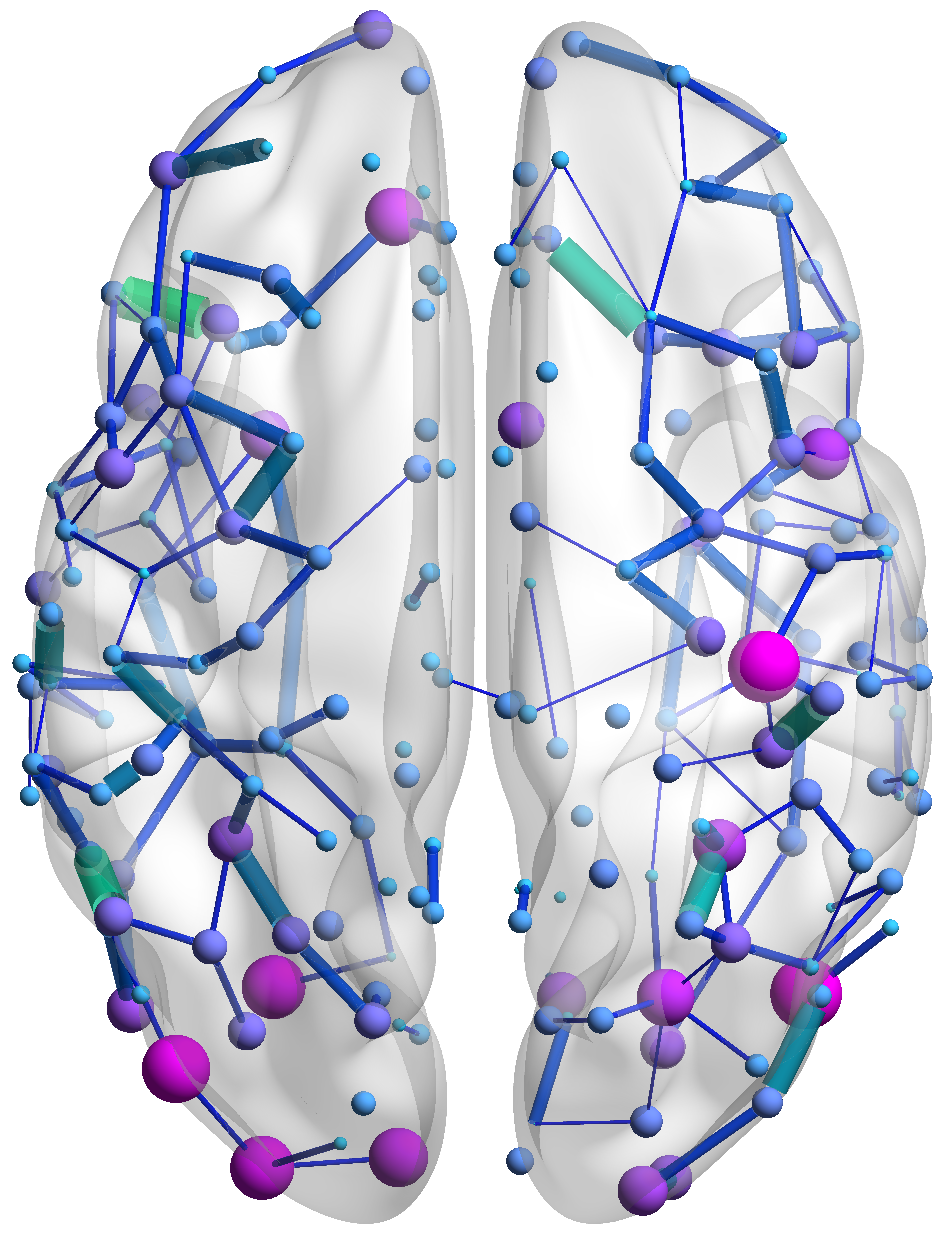}\hspace{+13pt}
    \includegraphics[width=0.017\textwidth]{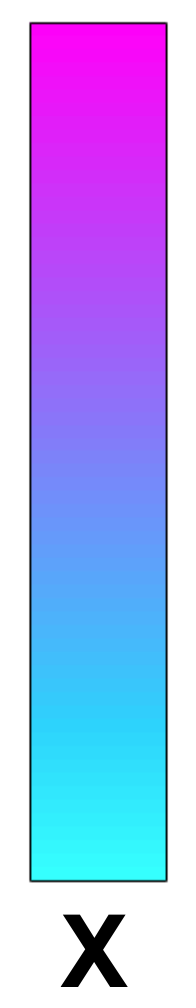}
    \caption{ROI attributions from $\operatorname{\textsc{Attr}}_{A}$ and $\operatorname{\textsc{Attr}}_{X}$. (Task order is the same as \cref{fig:attr_A_with_rand_colsum}). Edge color and width are based on task-averaged $\operatorname{\textsc{Attr}}_{A}\in \mathbb{R}^{200\times200}$, and node color and size are based on task and temporal-averaged $\operatorname{\textsc{Attr}}_{X}\in\mathbb{R}^{200}$. For visualization, only edges with highest attributions are shown (the resulting sparsity reduces to 0.009 from 0.196).
    }
    \label{fig:bnv_XAcombined}
\end{figure*}

\textbf{Spatial importance.}
We examine the connection importance between brain ROIs by looking at $\operatorname{\textsc{Attr}}_{A}$.
In particular, columns in $\operatorname{\textsc{Attr}}_{A}$ with higher average values are sender ROIs of high-contributing connections, which is what matters in the $\operatorname{GNN}$ operation.
We also explore why using random graph adjacency matrices (setting (v) in \cref{ssc:model_components}) can produce a similar result for length-256 inputs compared to using both SC-induced $A_i$ and $A_{i\_\text{adp}}$ (setting (i)). By examining $\operatorname{\textsc{Attr}}_{A}$ under both settings (\cref{fig:attr_A_with_rand_colsum}), we see that the column averages of $\operatorname{\textsc{Attr}}_{A}$ under these two settings are similar for almost all tasks, meaning the model can learn the important signal sending regions relatively well even without explicit structures.
We credit this ability primarily to multi-resolution inner cluster smoothing, as the performance drops notably without it (setting (vii)).
However, using ground truth SC not only gives us higher performance for shorter inputs but also provides the opportunity to interpret brain region connections better. We can directly use task-averaged $\operatorname{\textsc{Attr}}_{A}$ as the weighted adjacency matrix to plot edges between brain ROIs, just as in \cref{fig:bnv_XAcombined}. Important brain regions obtained from $\operatorname{\textsc{Attr}}_{A}$ mostly comply with the previous literature (see \cref{sssc:interp} for details).

In addition to $\operatorname{\textsc{Attr}}_{A}$, $\operatorname{\textsc{Attr}}_{X}$ can also provide insights on spatial importance when the attribution maps are aggregated along the temporal dimension. But it does so from another perspective: based on how the model takes in the inputs, larger $\operatorname{\textsc{Attr}}_{A}$ implies critical \textit{structural connections} between brain regions, meaning that information passing between those regions is deemed essential in classifying task states.
In contrast, larger $\operatorname{\textsc{Attr}}_{X}$ reveals regions or subnetworks that are sources of the important \textit{signals}: it does not matter if the signal activities propagate from one region to another. Instead, the signals themselves are crucial for differentiating between task states. 
We notice that signal-important ROIs are not necessarily the same as connection-important ROIs: top-ranked subnetworks for resting state are DefaultA and DefaultB by $\operatorname{\textsc{Attr}}_{A}$, and VisCent and DorsAttnA by $\operatorname{\textsc{Attr}}_{X}$; although they do coincide with each other for tasks like VMN. This disparity is reflected in \cref{fig:bnv_XAcombined} as edge and node differences. 
Another observation is that DYN and PVT have similar $\operatorname{\textsc{Attr}}_{A}$ patterns; both have a high attribution on connections originating from visual, control, and somatomotor systems. But when looking at $\operatorname{\textsc{Attr}}_{X}$, DYN and PVT are extreme opposites. For example, PVT has a very high $\operatorname{\textsc{Attr}}_{X}$ for a few ROIs in LH\_SomMotA, DorsAttnA\_TempOcc, and RH\_VisCent\_ExStr, while DYN has very low $\operatorname{\textsc{Attr}}_{X}$ for them. This suggests that the model uses these ROIs' activities to distinguish between the two tasks. Therefore, the attributions are not absolute but relative to what they are compared against. As a result, when identifying biomarkers with attribution, it is crucial to have \textit{contrasts}---for example, different tasks, different disease states, etc.

In \cref{fig:ig_task}, we plot the distribution of time-averaged and subnetwork-averaged (mapping 200 ROIs into 17 subnetworks) $\operatorname{\textsc{Attr}}_{X}$
during the VWM task. We can see the clear dominance of VisCent, DorsAttnA, and ContA subnetworks (numbered as 1, 5, 11), indicating signals from these regions are useful for the model to decide if the input is from the VWM task.
More informative than the rankings is the distribution itself: even though VisCent, DorsAttnA, and ContA ranked top 3 for both resting state and VWM for signal attributions, their relative importance and attribution distribution variances are drastically different. In a sense, the distribution can act as a task fingerprint based on brain signal states.

\textbf{Group, session, and region heterogeneity.} 
Average variances of attributions are very different across tasks, especially those of $\operatorname{\textsc{Attr}}_{X}$: VWM and DYN have much smaller attribution variances compared to other tasks. This can be caused by either task dynamics when certain tasks have more phase transitions and brain status changes, or/and group heterogeneity when individuals carry out specific tasks more differently than the others. We investigate this by examining three subjects that have multiple scan sessions for every task.
\begin{figure}[h]
    \centering
    \captionsetup{font=small}
    \includegraphics[width=0.235\textwidth]{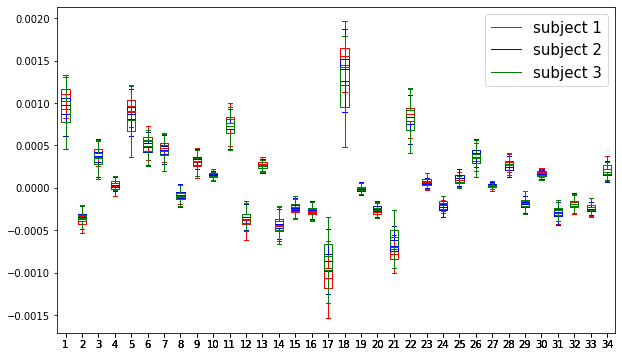}
    \includegraphics[width=0.235\textwidth]{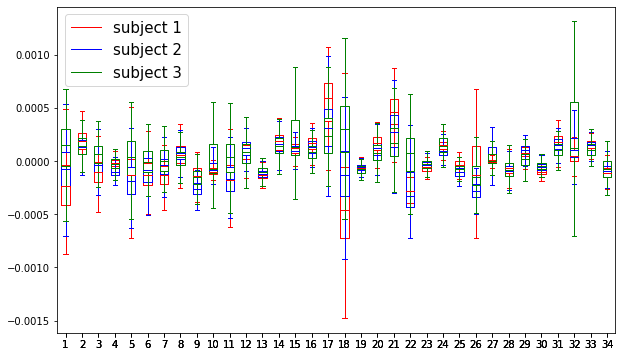}
    \caption{34 subnetworks' $\operatorname{\textsc{Attr}}_{X}$ distributions of 3 subjects performing the VWM task (left) and the MOD task (right). Outliers that go beyond \begin{small} $[Q1 - 1.5 \operatorname{IQR}, Q3 + 1.5 \operatorname{IQR}]$
    \end{small} are omitted. VWM has a much smaller average attribution variance than MOD.}
    \label{fig:hetero}
    \vspace{-0.3cm}
\end{figure}

We report the following findings: (1) Even only aggregating attributions over a single subject's sessions, attribution variances of the other four tasks are still larger than VWM and DYN. And these variance values are comparable to that of aggregating over many subjects. This means the large variances are not mainly due to group heterogeneity; rather, some tasks have more states than others. (2) There is still group heterogeneity apart from different task dynamics, and the group heterogeneity is also more evident for tasks with more dynamics (high attribution variances). We can see from \cref{fig:hetero} that attributions for VMM are much more concentrated and universal across subjects than that of MOD. (3) Flexibility of different subnetworks varies: subnetworks with small distribution IQR (interquartile range) of the same subject's different sessions are also more consistent across subjects. One example is that subnetwork 18 during the MOD task has both higher within-subject IQR and more significant across-subject differences than subnetwork 19. This indicates that for a particular task, some subnetworks are more individual and flexible (may activate differently across time), while others are more collective and fixed. In summary, we can find both critical regions that a particular task must rely on and regions that can characterize individual differences during tasks.

\subsection{Simulation study}
\begin{figure}
    \centering
    \captionsetup{font=small}
    \begin{subfigure}[b]{0.115\textwidth}
    \includegraphics[width=\textwidth]{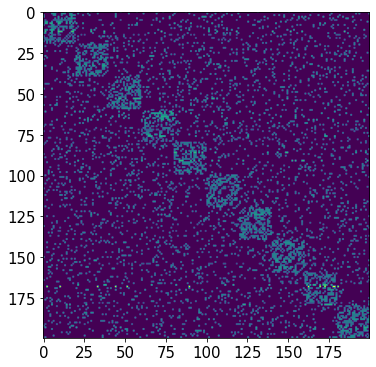}
    \caption{A}
    \label{fig:syn_A}
    \end{subfigure}
    \begin{subfigure}[b]{0.355\textwidth}
    \includegraphics[width=0.48\textwidth]{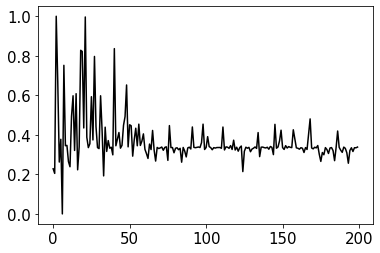}
    \includegraphics[width=0.48\textwidth]{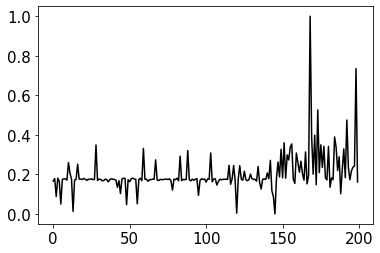} 
    \caption{Simulation (I) $\operatorname{Attr}_X$ of 200 nodes}
    \label{fig:syn_attr_X}
    \end{subfigure}
    \begin{subfigure}[b]{0.505\textwidth}
    \includegraphics[width=0.35\textwidth]{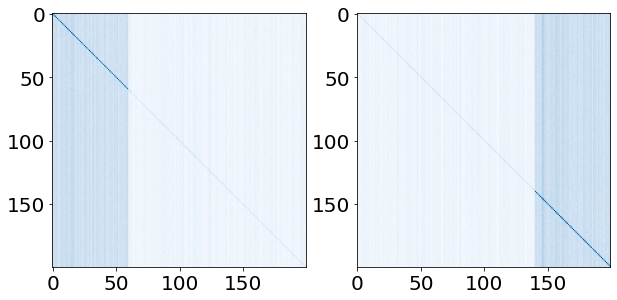}
    \includegraphics[width=0.62\textwidth]{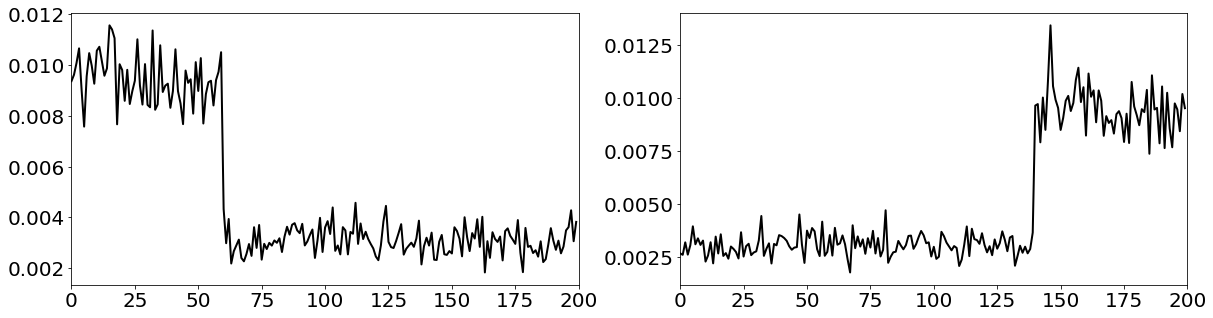}
    \caption{Simulation (I) $A_{\text{adp}}$ of 200 nodes}
    \label{fig:syn12_adp}
    \end{subfigure}
    \centering
    \begin{subfigure}[b]{0.5\textwidth}
    \includegraphics[width=0.37\textwidth]{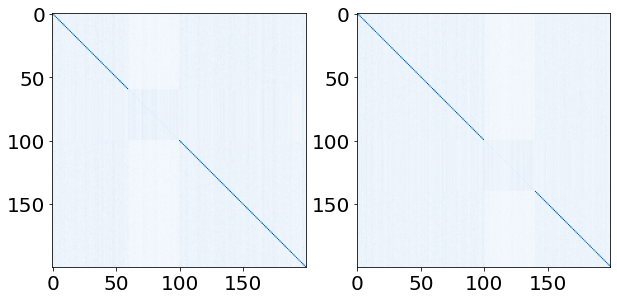}
    \includegraphics[trim={0 8pt 0 0}, clip,width=0.29\textwidth]{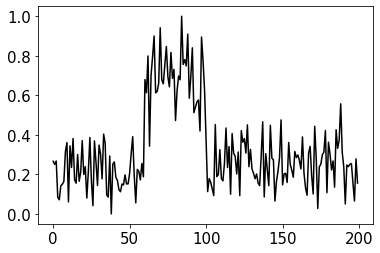}
    \includegraphics[trim={0 8pt 0 0}, clip, width=0.29\textwidth]{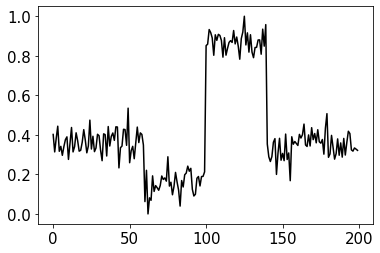}
    \caption{Simulation (II) $A_{\text{adp}}$ and $\operatorname{Attr}_A$ of 200 nodes}
    \label{fig:syn34_A}
    \end{subfigure}
    \caption{
    (a) A typical adjacency matrix for simulated graph signals.
    (b) Task averaged $\operatorname{Attr}_X$ of simulation (I). Attribution values are normalized.
    (c) Task averaged $A_{\text{adp}}$ of simulation (I) and its entry averages per column.
    (d) Task averaged $A_{\text{adp}}$ and task averaged $\operatorname{Attr}_A$ of simulation (II). Attribution values are normalized.}
    \label{fig:syn}
    \vspace{-0.3cm}
\end{figure}

To validate the results of our interpretations, we perform simulation studies with known ground truth.
All graphs are generated with SBM (stochastic block model) using the same community structure (200 nodes, 10 communities), but each graph has its own adjacency matrix. This generation process mimics brain structures in that samples share similar community structures but have distinct structural connectivities. Fig \ref{fig:syn_A} shows a typical adjacency matrix of a synthetic graph. All adjacency matrices are binary. Time-series on each node are then generated with code adapted from pytorch-gnn repository \footnote{\href{https://github.com/alelab-upenn/graph-neural-networks}{https://github.com/alelab-upenn/graph-neural-networks}}. In particular, the value at each time step of each node is a small temporal Gaussian random noise plus signals from neighbors' (a small spatial Gaussian noise is added to the adjacency matrix) previous step.

\textbf{Simulation (I)} We create two classes for this simulation. In class one, only the first three communities (nodes 1--60) generate small temporal noises, and other nodes are only affected by neighbors. In class two, only the last three communities (nodes 141--200) generate small temporal noises, and other nodes are only affected by neighbors. We visualize the task aggregated  $\operatorname{Attr}_X$ and $A_{\text{adp}}$ and in \cref{fig:syn_attr_X,fig:syn12_adp}. The signals are characterized well in $\operatorname{Attr}_X$. For the generated series, signals are more important in node 1--60 for class 1 and 141--200 for class 2: $A_{\text{adp}}$ finds this pattern and helps propagate signals in these regions better.
We notice that $\operatorname{Attr}_A$ is mostly random, with no apparent patterns. This is consistent with the graph signal generation: when aggregating information from neighbors, all connected edges are weighted the same (binary); thus, the connections do not affect generated signals. We perform the following study to understand the opposite effect.

\textbf{Simulation (II)} We again create two classes for the simulation: in class one, connections from nodes 61--100 are strengthened; in class two, connections from nodes 101--140 are strengthened. The weights of strengthened edges are increased from 1 to 5 during signal generation. However, the model still takes in binary adjacency matrices as inputs (processed as mentioned in \cref{sc:preliminaries} before feeding to the model). We visualize the task aggregated $A_{\text{adp}}$ and $\operatorname{Attr}_A$ in \cref{fig:syn34_A}. This time the connection differences are reflected in $\operatorname{Attr}_A$.
Signals in node 61--100 for class 1 or 101--140 for class 2 are less important because stronger connections can send these signals out: this results in smaller values for corresponding columns in $A_{\text{adp}}$. Combined with the previous simulation results, this suggests that strong signal sending regions 
or regions with weak connections that are over-reflected in the graph adjacency matrix 
tend to have higher $A_{\text{adp}}$ values. In other words, $A_{\text{adp}}$ complements both signals and connections to encode latent dynamics, while attributions obtained from IG are better at interpreting the modalities separately.

\section{Conclusions}
This paper proposes ReBraiD, a high-performing and efficient graph neural network model that embeds both structural and dynamic functional signals for a more comprehensive representation of brain dynamics. To better capture latent structures, we propose sample-level adjacency matrix learning and multi-resolution inner cluster smoothing. Apart from quantitative results showing ReBraiD's superiority in representing brain activities, we also leverage integrated gradients to attribute and interpret the importance of both spatial brain regions and temporal keyframes. The attribution also reveals heterogeneities among brain regions (or subnetworks), tasks, and individuals. These findings can potentially reveal new neural basis, biomarkers of tasks or brain disorders when combined with behavioral metrics. They can also enable more fine-grained temporal analysis around keyframes when combined with other imaging techniques and extend to different scientific domains with sample (subject) heterogeneity.
\begin{acks}
This project was  partially supported by funding from the National Science Foundation under grant IIS-1817046.
\end{acks}

\bibliographystyle{ACM-Reference-Format}
\bibliography{sample-base}


\begin{thebibliography}{43}


\ifx \showCODEN    \undefined \def \showCODEN     #1{\unskip}     \fi
\ifx \showDOI      \undefined \def \showDOI       #1{#1}\fi
\ifx \showISBNx    \undefined \def \showISBNx     #1{\unskip}     \fi
\ifx \showISBNxiii \undefined \def \showISBNxiii  #1{\unskip}     \fi
\ifx \showISSN     \undefined \def \showISSN      #1{\unskip}     \fi
\ifx \showLCCN     \undefined \def \showLCCN      #1{\unskip}     \fi
\ifx \shownote     \undefined \def \shownote      #1{#1}          \fi
\ifx \showarticletitle \undefined \def \showarticletitle #1{#1}   \fi
\ifx \showURL      \undefined \def \showURL       {\relax}        \fi
\providecommand\bibfield[2]{#2}
\providecommand\bibinfo[2]{#2}
\providecommand\natexlab[1]{#1}
\providecommand\showeprint[2][]{arXiv:#2}

\bibitem[Bassett and Sporns(2017)]%
        {bassett2017network}
\bibfield{author}{\bibinfo{person}{Danielle~S Bassett} {and}
  \bibinfo{person}{Olaf Sporns}.} \bibinfo{year}{2017}\natexlab{}.
\newblock \showarticletitle{Network neuroscience}.
\newblock \bibinfo{journal}{\emph{Nature Neuroscience}} \bibinfo{volume}{20},
  \bibinfo{number}{3} (\bibinfo{year}{2017}), \bibinfo{pages}{353--364}.
\newblock


\bibitem[Brody et~al\mbox{.}(2021)]%
        {brody2021attentive}
\bibfield{author}{\bibinfo{person}{Shaked Brody}, \bibinfo{person}{Uri Alon},
  {and} \bibinfo{person}{Eran Yahav}.} \bibinfo{year}{2021}\natexlab{}.
\newblock \bibinfo{title}{How Attentive are Graph Attention Networks?}
\newblock
\newblock
\showeprint[arxiv]{2105.14491}~[cs.LG]


\bibitem[Carlson et~al\mbox{.}(2012)]%
        {carlson2012nonconscious}
\bibfield{author}{\bibinfo{person}{Joshua~M Carlson}, \bibinfo{person}{Felix
  Beacher}, \bibinfo{person}{Karen~S Reinke}, \bibinfo{person}{Reza Habib},
  \bibinfo{person}{Eddie Harmon-Jones}, \bibinfo{person}{Lilianne~R
  Mujica-Parodi}, {and} \bibinfo{person}{Greg Hajcak}.}
  \bibinfo{year}{2012}\natexlab{}.
\newblock \showarticletitle{Nonconscious attention bias to threat is correlated
  with anterior cingulate cortex gray matter volume: a voxel-based morphometry
  result and replication}.
\newblock \bibinfo{journal}{\emph{Neuroimage}} \bibinfo{volume}{59},
  \bibinfo{number}{2} (\bibinfo{year}{2012}), \bibinfo{pages}{1713--1718}.
\newblock


\bibitem[Carlson et~al\mbox{.}(2013)]%
        {carlson2013functional}
\bibfield{author}{\bibinfo{person}{Joshua~M Carlson}, \bibinfo{person}{Jiook
  Cha}, {and} \bibinfo{person}{Lilianne~R Mujica-Parodi}.}
  \bibinfo{year}{2013}\natexlab{}.
\newblock \showarticletitle{Functional and structural amygdala--anterior
  cingulate connectivity correlates with attentional bias to masked fearful
  faces}.
\newblock \bibinfo{journal}{\emph{Cortex}} \bibinfo{volume}{49},
  \bibinfo{number}{9} (\bibinfo{year}{2013}), \bibinfo{pages}{2595--2600}.
\newblock


\bibitem[Drummond et~al\mbox{.}(2005)]%
        {drummond2005neural}
\bibfield{author}{\bibinfo{person}{Sean~PA Drummond}, \bibinfo{person}{Amanda
  Bischoff-Grethe}, \bibinfo{person}{David~F Dinges}, \bibinfo{person}{Liat
  Ayalon}, \bibinfo{person}{Sara~C Mednick}, {and} \bibinfo{person}{MJ Meloy}.}
  \bibinfo{year}{2005}\natexlab{}.
\newblock \showarticletitle{The neural basis of the psychomotor vigilance
  task}.
\newblock \bibinfo{journal}{\emph{Sleep}} \bibinfo{volume}{28},
  \bibinfo{number}{9} (\bibinfo{year}{2005}), \bibinfo{pages}{1059--1068}.
\newblock


\bibitem[Friedrich and Friederici(2013)]%
        {friedrich2013mathematical}
\bibfield{author}{\bibinfo{person}{Roland~M Friedrich} {and}
  \bibinfo{person}{Angela~D Friederici}.} \bibinfo{year}{2013}\natexlab{}.
\newblock \showarticletitle{Mathematical logic in the human brain: semantics}.
\newblock \bibinfo{journal}{\emph{PLoS One}} \bibinfo{volume}{8},
  \bibinfo{number}{1} (\bibinfo{year}{2013}), \bibinfo{pages}{e53699}.
\newblock


\bibitem[Grabner et~al\mbox{.}(2011)]%
        {grabner2011brain}
\bibfield{author}{\bibinfo{person}{Roland~H Grabner}, \bibinfo{person}{Gernot
  Reishofer}, \bibinfo{person}{Karl Koschutnig}, {and} \bibinfo{person}{Franz
  Ebner}.} \bibinfo{year}{2011}\natexlab{}.
\newblock \showarticletitle{Brain correlates of mathematical competence in
  processing mathematical representations}.
\newblock \bibinfo{journal}{\emph{Frontiers in Human Neuroscience}}
  \bibinfo{volume}{5} (\bibinfo{year}{2011}), \bibinfo{pages}{130}.
\newblock


\bibitem[Hamilton et~al\mbox{.}(2017)]%
        {hamilton2017inductive}
\bibfield{author}{\bibinfo{person}{William~L Hamilton}, \bibinfo{person}{Rex
  Ying}, {and} \bibinfo{person}{Jure Leskovec}.}
  \bibinfo{year}{2017}\natexlab{}.
\newblock \showarticletitle{Inductive representation learning on large graphs}.
  In \bibinfo{booktitle}{\emph{Proceedings of the 31st International Conference
  on Neural Information Processing Systems}}. \bibinfo{pages}{1025--1035}.
\newblock


\bibitem[Kim and Ye(2020)]%
        {10.3389/fnins.2020.00630}
\bibfield{author}{\bibinfo{person}{Byung-Hoon Kim} {and}
  \bibinfo{person}{Jong~Chul Ye}.} \bibinfo{year}{2020}\natexlab{}.
\newblock \showarticletitle{Understanding Graph Isomorphism Network for rs-fMRI
  Functional Connectivity Analysis}.
\newblock \bibinfo{journal}{\emph{Frontiers in Neuroscience}}
  \bibinfo{volume}{14} (\bibinfo{year}{2020}), \bibinfo{pages}{630}.
\newblock
\showISSN{1662-453X}
\urldef\tempurl%
\url{https://doi.org/10.3389/fnins.2020.00630}
\showDOI{\tempurl}


\bibitem[Kim et~al\mbox{.}(2016)]%
        {kim2016anterior}
\bibfield{author}{\bibinfo{person}{Jangjin Kim}, \bibinfo{person}{Edward~A
  Wasserman}, \bibinfo{person}{Leyre Castro}, {and} \bibinfo{person}{John~H
  Freeman}.} \bibinfo{year}{2016}\natexlab{}.
\newblock \showarticletitle{Anterior cingulate cortex inactivation impairs
  rodent visual selective attention and prospective memory.}
\newblock \bibinfo{journal}{\emph{Behavioral Neuroscience}}
  \bibinfo{volume}{130}, \bibinfo{number}{1} (\bibinfo{year}{2016}),
  \bibinfo{pages}{75}.
\newblock


\bibitem[Kipf and Welling(2017)]%
        {kipf2017semisupervised}
\bibfield{author}{\bibinfo{person}{Thomas~N. Kipf} {and} \bibinfo{person}{Max
  Welling}.} \bibinfo{year}{2017}\natexlab{}.
\newblock \showarticletitle{Semi-Supervised Classification with Graph
  Convolutional Networks}. In \bibinfo{booktitle}{\emph{International
  Conference on Learning Representations (ICLR)}}.
\newblock


\bibitem[Lauharatanahirun et~al\mbox{.}(2020)]%
        {lauharatanahirun2020flexibility}
\bibfield{author}{\bibinfo{person}{Nina Lauharatanahirun},
  \bibinfo{person}{Kanika Bansal}, \bibinfo{person}{Steven~M Thurman},
  \bibinfo{person}{Jean~M Vettel}, \bibinfo{person}{Barry Giesbrecht},
  \bibinfo{person}{Scott Grafton}, \bibinfo{person}{James~C Elliott},
  \bibinfo{person}{Erin Flynn-Evans}, \bibinfo{person}{Emily Falk}, {and}
  \bibinfo{person}{Javier~O Garcia}.} \bibinfo{year}{2020}\natexlab{}.
\newblock \showarticletitle{Flexibility of brain regions during working memory
  curtails cognitive consequences to lack of sleep}.
\newblock \bibinfo{journal}{\emph{arXiv preprint arXiv:2009.07233}}
  (\bibinfo{year}{2020}).
\newblock


\bibitem[Lea et~al\mbox{.}(2016)]%
        {10.1007/978-3-319-49409-8_7}
\bibfield{author}{\bibinfo{person}{Colin Lea}, \bibinfo{person}{Ren{\'e}
  Vidal}, \bibinfo{person}{Austin Reiter}, {and} \bibinfo{person}{Gregory~D.
  Hager}.} \bibinfo{year}{2016}\natexlab{}.
\newblock \showarticletitle{Temporal Convolutional Networks: A Unified Approach
  to Action Segmentation}. In \bibinfo{booktitle}{\emph{Computer Vision -- ECCV
  2016 Workshops}}, \bibfield{editor}{\bibinfo{person}{Gang Hua} {and}
  \bibinfo{person}{Herv{\'e} J{\'e}gou}} (Eds.). \bibinfo{publisher}{Springer
  International Publishing}, \bibinfo{address}{Cham}, \bibinfo{pages}{47--54}.
\newblock
\showISBNx{978-3-319-49409-8}


\bibitem[Leech and Sharp(2014)]%
        {leech2014role}
\bibfield{author}{\bibinfo{person}{Robert Leech} {and} \bibinfo{person}{David~J
  Sharp}.} \bibinfo{year}{2014}\natexlab{}.
\newblock \showarticletitle{The role of the posterior cingulate cortex in
  cognition and disease}.
\newblock \bibinfo{journal}{\emph{Brain}} \bibinfo{volume}{137},
  \bibinfo{number}{1} (\bibinfo{year}{2014}), \bibinfo{pages}{12--32}.
\newblock


\bibitem[Li et~al\mbox{.}(2019)]%
        {li2019modeling}
\bibfield{author}{\bibinfo{person}{Lingge Li}, \bibinfo{person}{Dustin Pluta},
  \bibinfo{person}{Babak Shahbaba}, \bibinfo{person}{Norbert Fortin},
  \bibinfo{person}{Hernando Ombao}, {and} \bibinfo{person}{Pierre Baldi}.}
  \bibinfo{year}{2019}\natexlab{}.
\newblock \showarticletitle{Modeling dynamic functional connectivity with
  latent factor Gaussian processes}.
\newblock \bibinfo{journal}{\emph{Advances in Neural Information Processing
  Systems}}  \bibinfo{volume}{32} (\bibinfo{year}{2019}),
  \bibinfo{pages}{8263--8273}.
\newblock


\bibitem[Li et~al\mbox{.}(2020)]%
        {Li2020PoolingRG}
\bibfield{author}{\bibinfo{person}{Xiaoxiao Li}, \bibinfo{person}{Yuan Zhou},
  \bibinfo{person}{Nicha~C. Dvornek}, \bibinfo{person}{Muhan Zhang},
  \bibinfo{person}{Juntang Zhuang}, \bibinfo{person}{Pamela Ventola}, {and}
  \bibinfo{person}{James~S. Duncan}.} \bibinfo{year}{2020}\natexlab{}.
\newblock \showarticletitle{Pooling Regularized Graph Neural Network for fMRI
  Biomarker Analysis}.
\newblock \bibinfo{journal}{\emph{Medical Image Computing and Computer-assisted
  Intervention (MICCAI)}}  \bibinfo{volume}{12267} (\bibinfo{year}{2020}),
  \bibinfo{pages}{625--635}.
\newblock


\bibitem[Liu et~al\mbox{.}(2020)]%
        {liu2020disentangling}
\bibfield{author}{\bibinfo{person}{Ziyu Liu}, \bibinfo{person}{Hongwen Zhang},
  \bibinfo{person}{Zhenghao Chen}, \bibinfo{person}{Zhiyong Wang}, {and}
  \bibinfo{person}{Wanli Ouyang}.} \bibinfo{year}{2020}\natexlab{}.
\newblock \showarticletitle{Disentangling and unifying graph convolutions for
  skeleton-based action recognition}. In \bibinfo{booktitle}{\emph{Proceedings
  of the IEEE/CVF Conference on Computer Vision and Pattern Recognition}}.
  \bibinfo{pages}{143--152}.
\newblock


\bibitem[Loh et~al\mbox{.}(2004)]%
        {loh2004validity}
\bibfield{author}{\bibinfo{person}{Sylvia Loh}, \bibinfo{person}{Nicole
  Lamond}, \bibinfo{person}{Jill Dorrian}, \bibinfo{person}{Gregory Roach},
  {and} \bibinfo{person}{Drew Dawson}.} \bibinfo{year}{2004}\natexlab{}.
\newblock \showarticletitle{The validity of psychomotor vigilance tasks of less
  than 10-minute duration}.
\newblock \bibinfo{journal}{\emph{Behavior Research Methods, Instruments, \&
  Computers}} \bibinfo{volume}{36}, \bibinfo{number}{2} (\bibinfo{year}{2004}),
  \bibinfo{pages}{339--346}.
\newblock


\bibitem[Luck and Vogel(1997)]%
        {luck1997capacity}
\bibfield{author}{\bibinfo{person}{Steven~J Luck} {and}
  \bibinfo{person}{Edward~K Vogel}.} \bibinfo{year}{1997}\natexlab{}.
\newblock \showarticletitle{The capacity of visual working memory for features
  and conjunctions}.
\newblock \bibinfo{journal}{\emph{Nature}} \bibinfo{volume}{390},
  \bibinfo{number}{6657} (\bibinfo{year}{1997}), \bibinfo{pages}{279--281}.
\newblock


\bibitem[Mattarella-Micke et~al\mbox{.}(2011)]%
        {mattarella2011choke}
\bibfield{author}{\bibinfo{person}{Andrew Mattarella-Micke},
  \bibinfo{person}{Jill Mateo}, \bibinfo{person}{Megan~N Kozak},
  \bibinfo{person}{Katherine Foster}, {and} \bibinfo{person}{Sian~L Beilock}.}
  \bibinfo{year}{2011}\natexlab{}.
\newblock \showarticletitle{Choke or thrive? The relation between salivary
  cortisol and math performance depends on individual differences in working
  memory and math-anxiety.}
\newblock \bibinfo{journal}{\emph{Emotion}} \bibinfo{volume}{11},
  \bibinfo{number}{4} (\bibinfo{year}{2011}), \bibinfo{pages}{1000}.
\newblock


\bibitem[Noman et~al\mbox{.}(2021)]%
        {noman2021graph}
\bibfield{author}{\bibinfo{person}{Fuad Noman}, \bibinfo{person}{Chee-Ming
  Ting}, \bibinfo{person}{Hakmook Kang}, \bibinfo{person}{Raphael C.~W. Phan},
  \bibinfo{person}{Brian~D. Boyd}, \bibinfo{person}{Warren~D. Taylor}, {and}
  \bibinfo{person}{Hernando Ombao}.} \bibinfo{year}{2021}\natexlab{}.
\newblock \bibinfo{title}{Graph Autoencoders for Embedding Learning in Brain
  Networks and Major Depressive Disorder Identification}.
\newblock
\newblock
\showeprint[arxiv]{2107.12838}~[q-bio.NC]


\bibitem[Oord et~al\mbox{.}(2016)]%
        {oord2016conditional}
\bibfield{author}{\bibinfo{person}{A{\"a}ron van~den Oord},
  \bibinfo{person}{Nal Kalchbrenner}, \bibinfo{person}{Oriol Vinyals},
  \bibinfo{person}{Lasse Espeholt}, \bibinfo{person}{Alex Graves}, {and}
  \bibinfo{person}{Koray Kavukcuoglu}.} \bibinfo{year}{2016}\natexlab{}.
\newblock \showarticletitle{Conditional image generation with PixelCNN
  decoders}. In \bibinfo{booktitle}{\emph{Proceedings of the 30th International
  Conference on Neural Information Processing Systems}}.
  \bibinfo{pages}{4797--4805}.
\newblock


\bibitem[Raichle(2015)]%
        {raichle2015brain}
\bibfield{author}{\bibinfo{person}{Marcus~E Raichle}.}
  \bibinfo{year}{2015}\natexlab{}.
\newblock \showarticletitle{The brain's default mode network}.
\newblock \bibinfo{journal}{\emph{Annual Review of Neuroscience}}
  \bibinfo{volume}{38} (\bibinfo{year}{2015}), \bibinfo{pages}{433--447}.
\newblock


\bibitem[Ruiz et~al\mbox{.}(2020)]%
        {ruiz2020gated}
\bibfield{author}{\bibinfo{person}{Luana Ruiz}, \bibinfo{person}{Fernando
  Gama}, {and} \bibinfo{person}{Alejandro Ribeiro}.}
  \bibinfo{year}{2020}\natexlab{}.
\newblock \showarticletitle{Gated graph recurrent neural networks}.
\newblock \bibinfo{journal}{\emph{IEEE Transactions on Signal Processing}}
  \bibinfo{volume}{68} (\bibinfo{year}{2020}), \bibinfo{pages}{6303--6318}.
\newblock


\bibitem[Schaefer et~al\mbox{.}(2018)]%
        {schaefer2018local}
\bibfield{author}{\bibinfo{person}{Alexander Schaefer}, \bibinfo{person}{Ru
  Kong}, \bibinfo{person}{Evan~M Gordon}, \bibinfo{person}{Timothy~O Laumann},
  \bibinfo{person}{Xi-Nian Zuo}, \bibinfo{person}{Avram~J Holmes},
  \bibinfo{person}{Simon~B Eickhoff}, {and} \bibinfo{person}{BT~Thomas Yeo}.}
  \bibinfo{year}{2018}\natexlab{}.
\newblock \showarticletitle{Local-global parcellation of the human cerebral
  cortex from intrinsic functional connectivity MRI}.
\newblock \bibinfo{journal}{\emph{Cerebral Cortex}} \bibinfo{volume}{28},
  \bibinfo{number}{9} (\bibinfo{year}{2018}), \bibinfo{pages}{3095--3114}.
\newblock


\bibitem[Selvaraju et~al\mbox{.}(2017)]%
        {selvaraju2017grad}
\bibfield{author}{\bibinfo{person}{Ramprasaath~R Selvaraju},
  \bibinfo{person}{Michael Cogswell}, \bibinfo{person}{Abhishek Das},
  \bibinfo{person}{Ramakrishna Vedantam}, \bibinfo{person}{Devi Parikh}, {and}
  \bibinfo{person}{Dhruv Batra}.} \bibinfo{year}{2017}\natexlab{}.
\newblock \showarticletitle{Grad-cam: Visual explanations from deep networks
  via gradient-based localization}. In \bibinfo{booktitle}{\emph{Proceedings of
  the IEEE International Conference on Computer Vision}}.
  \bibinfo{pages}{618--626}.
\newblock


\bibitem[Seo et~al\mbox{.}(2018)]%
        {seo2018structured}
\bibfield{author}{\bibinfo{person}{Youngjoo Seo}, \bibinfo{person}{Micha{\"e}l
  Defferrard}, \bibinfo{person}{Pierre Vandergheynst}, {and}
  \bibinfo{person}{Xavier Bresson}.} \bibinfo{year}{2018}\natexlab{}.
\newblock \showarticletitle{Structured sequence modeling with graph
  convolutional recurrent networks}. In \bibinfo{booktitle}{\emph{International
  Conference on Neural Information Processing}}. Springer,
  \bibinfo{pages}{362--373}.
\newblock


\bibitem[Shi et~al\mbox{.}(2021)]%
        {shi2021masked}
\bibfield{author}{\bibinfo{person}{Yunsheng Shi}, \bibinfo{person}{Zhengjie
  Huang}, \bibinfo{person}{Shikun Feng}, \bibinfo{person}{Hui Zhong},
  \bibinfo{person}{Wenjin Wang}, {and} \bibinfo{person}{Yu Sun}.}
  \bibinfo{year}{2021}\natexlab{}.
\newblock \bibinfo{title}{Masked Label Prediction: Unified Message Passing
  Model for Semi-Supervised Classification}.
\newblock
\newblock
\showeprint[arxiv]{2009.03509}~[cs.LG]


\bibitem[Sipos et~al\mbox{.}(2014)]%
        {sipos2014postdeployment}
\bibfield{author}{\bibinfo{person}{Maurice~L Sipos}, \bibinfo{person}{Yair
  Bar-Haim}, \bibinfo{person}{Rany Abend}, \bibinfo{person}{Amy~B Adler}, {and}
  \bibinfo{person}{Paul~D Bliese}.} \bibinfo{year}{2014}\natexlab{}.
\newblock \showarticletitle{Postdeployment threat-related attention bias
  interacts with combat exposure to account for PTSD and anxiety symptoms in
  soldiers}.
\newblock \bibinfo{journal}{\emph{Depression and Anxiety}}
  \bibinfo{volume}{31}, \bibinfo{number}{2} (\bibinfo{year}{2014}),
  \bibinfo{pages}{124--129}.
\newblock


\bibitem[Song et~al\mbox{.}(2020)]%
        {song2020spatial}
\bibfield{author}{\bibinfo{person}{Chao Song}, \bibinfo{person}{Youfang Lin},
  \bibinfo{person}{Shengnan Guo}, {and} \bibinfo{person}{Huaiyu Wan}.}
  \bibinfo{year}{2020}\natexlab{}.
\newblock \showarticletitle{Spatial-temporal synchronous graph convolutional
  networks: A new framework for spatial-temporal network data forecasting}. In
  \bibinfo{booktitle}{\emph{Proceedings of the AAAI Conference on Artificial
  Intelligence}}, Vol.~\bibinfo{volume}{34}. \bibinfo{pages}{914--921}.
\newblock


\bibitem[Sundararajan et~al\mbox{.}(2017)]%
        {sundararajan2017axiomatic}
\bibfield{author}{\bibinfo{person}{Mukund Sundararajan}, \bibinfo{person}{Ankur
  Taly}, {and} \bibinfo{person}{Qiqi Yan}.} \bibinfo{year}{2017}\natexlab{}.
\newblock \showarticletitle{Axiomatic attribution for deep networks}. In
  \bibinfo{booktitle}{\emph{International Conference on Machine Learning}}.
  PMLR, \bibinfo{pages}{3319--3328}.
\newblock


\bibitem[Thomas~Yeo et~al\mbox{.}(2011)]%
        {thomas2011organization}
\bibfield{author}{\bibinfo{person}{BT Thomas~Yeo}, \bibinfo{person}{Fenna~M
  Krienen}, \bibinfo{person}{Jorge Sepulcre}, \bibinfo{person}{Mert~R Sabuncu},
  \bibinfo{person}{Danial Lashkari}, \bibinfo{person}{Marisa Hollinshead},
  \bibinfo{person}{Joshua~L Roffman}, \bibinfo{person}{Jordan~W Smoller},
  \bibinfo{person}{Lilla Z{\"o}llei}, \bibinfo{person}{Jonathan~R Polimeni},
  {et~al\mbox{.}}} \bibinfo{year}{2011}\natexlab{}.
\newblock \showarticletitle{The organization of the human cerebral cortex
  estimated by intrinsic functional connectivity}.
\newblock \bibinfo{journal}{\emph{Journal of Neurophysiology}}
  \bibinfo{volume}{106}, \bibinfo{number}{3} (\bibinfo{year}{2011}),
  \bibinfo{pages}{1125--1165}.
\newblock


\bibitem[Tishby and Zaslavsky(2015)]%
        {tishby2015deep}
\bibfield{author}{\bibinfo{person}{Naftali Tishby} {and} \bibinfo{person}{Noga
  Zaslavsky}.} \bibinfo{year}{2015}\natexlab{}.
\newblock \showarticletitle{Deep learning and the information bottleneck
  principle}. In \bibinfo{booktitle}{\emph{2015 IEEE Information Theory
  Workshop (ITW)}}. IEEE, \bibinfo{pages}{1--5}.
\newblock


\bibitem[Todd and Marois(2004)]%
        {todd2004capacity}
\bibfield{author}{\bibinfo{person}{J~Jay Todd} {and} \bibinfo{person}{Ren{\'e}
  Marois}.} \bibinfo{year}{2004}\natexlab{}.
\newblock \showarticletitle{Capacity limit of visual short-term memory in human
  posterior parietal cortex}.
\newblock \bibinfo{journal}{\emph{Nature}} \bibinfo{volume}{428},
  \bibinfo{number}{6984} (\bibinfo{year}{2004}), \bibinfo{pages}{751--754}.
\newblock


\bibitem[{van den Oord} et~al\mbox{.}(2016)]%
        {oord2016wavenet}
\bibfield{author}{\bibinfo{person}{Aäron {van den Oord}},
  \bibinfo{person}{Sander Dieleman}, \bibinfo{person}{Heiga Zen},
  \bibinfo{person}{Karen Simonyan}, \bibinfo{person}{Oriol Vinyals},
  \bibinfo{person}{Alex Graves}, \bibinfo{person}{Nal Kalchbrenner},
  \bibinfo{person}{Andrew Senior}, {and} \bibinfo{person}{Koray Kavukcuoglu}.}
  \bibinfo{year}{2016}\natexlab{}.
\newblock \showarticletitle{{WaveNet: A Generative Model for Raw Audio}}. In
  \bibinfo{booktitle}{\emph{Proc. 9th ISCA Workshop on Speech Synthesis
  Workshop (SSW 9)}}. \bibinfo{pages}{125}.
\newblock


\bibitem[Wiltschko et~al\mbox{.}(2020)]%
        {wiltschko2020evaluating}
\bibfield{author}{\bibinfo{person}{Alexander~B Wiltschko},
  \bibinfo{person}{Benjamin Sanchez-Lengeling}, \bibinfo{person}{Brian Lee},
  \bibinfo{person}{Emily Reif}, \bibinfo{person}{Jennifer Wei},
  \bibinfo{person}{Kevin~James McCloskey}, \bibinfo{person}{Lucy Colwell},
  \bibinfo{person}{Wesley Qian}, {and} \bibinfo{person}{Yiliu Wang}.}
  \bibinfo{year}{2020}\natexlab{}.
\newblock \showarticletitle{Evaluating Attribution for Graph Neural Networks}.
\newblock \bibinfo{journal}{\emph{Google Research}} (\bibinfo{year}{2020}).
\newblock


\bibitem[Wu et~al\mbox{.}(2019b)]%
        {wu2019simplifying}
\bibfield{author}{\bibinfo{person}{Felix Wu}, \bibinfo{person}{Amauri Souza},
  \bibinfo{person}{Tianyi Zhang}, \bibinfo{person}{Christopher Fifty},
  \bibinfo{person}{Tao Yu}, {and} \bibinfo{person}{Kilian Weinberger}.}
  \bibinfo{year}{2019}\natexlab{b}.
\newblock \showarticletitle{Simplifying graph convolutional networks}. In
  \bibinfo{booktitle}{\emph{International Conference on Machine Learning}}.
  PMLR, \bibinfo{pages}{6861--6871}.
\newblock


\bibitem[Wu et~al\mbox{.}(2019a)]%
        {wu2019graph}
\bibfield{author}{\bibinfo{person}{Zonghan Wu}, \bibinfo{person}{Shirui Pan},
  \bibinfo{person}{Guodong Long}, \bibinfo{person}{Jing Jiang}, {and}
  \bibinfo{person}{Chengqi Zhang}.} \bibinfo{year}{2019}\natexlab{a}.
\newblock \showarticletitle{Graph wavenet for deep spatial-temporal graph
  modeling}.
\newblock \bibinfo{journal}{\emph{International Joint Conferences on Artificial
  Intelligence (IJCAI)}} (\bibinfo{year}{2019}).
\newblock


\bibitem[Xu et~al\mbox{.}(2018)]%
        {xu2018powerful}
\bibfield{author}{\bibinfo{person}{Keyulu Xu}, \bibinfo{person}{Weihua Hu},
  \bibinfo{person}{Jure Leskovec}, {and} \bibinfo{person}{Stefanie Jegelka}.}
  \bibinfo{year}{2018}\natexlab{}.
\newblock \showarticletitle{How Powerful are Graph Neural Networks?}. In
  \bibinfo{booktitle}{\emph{International Conference on Learning
  Representations}}.
\newblock


\bibitem[Yantis et~al\mbox{.}(2002)]%
        {yantis2002transient}
\bibfield{author}{\bibinfo{person}{Steven Yantis}, \bibinfo{person}{Jens
  Schwarzbach}, \bibinfo{person}{John~T Serences}, \bibinfo{person}{Robert~L
  Carlson}, \bibinfo{person}{Michael~A Steinmetz}, \bibinfo{person}{James~J
  Pekar}, {and} \bibinfo{person}{Susan~M Courtney}.}
  \bibinfo{year}{2002}\natexlab{}.
\newblock \showarticletitle{Transient neural activity in human parietal cortex
  during spatial attention shifts}.
\newblock \bibinfo{journal}{\emph{Nature Neuroscience}} \bibinfo{volume}{5},
  \bibinfo{number}{10} (\bibinfo{year}{2002}), \bibinfo{pages}{995--1002}.
\newblock


\bibitem[Ying et~al\mbox{.}(2018)]%
        {ying2018hierarchical}
\bibfield{author}{\bibinfo{person}{Rex Ying}, \bibinfo{person}{Jiaxuan You},
  \bibinfo{person}{Christopher Morris}, \bibinfo{person}{Xiang Ren},
  \bibinfo{person}{William~L Hamilton}, {and} \bibinfo{person}{Jure Leskovec}.}
  \bibinfo{year}{2018}\natexlab{}.
\newblock \showarticletitle{Hierarchical graph representation learning with
  differentiable pooling}. In \bibinfo{booktitle}{\emph{Proceedings of the 32nd
  International Conference on Neural Information Processing Systems}}.
  \bibinfo{pages}{4805--4815}.
\newblock


\bibitem[Zerveas et~al\mbox{.}(2021)]%
        {zerveas2021transformer}
\bibfield{author}{\bibinfo{person}{George Zerveas}, \bibinfo{person}{Srideepika
  Jayaraman}, \bibinfo{person}{Dhaval Patel}, \bibinfo{person}{Anuradha
  Bhamidipaty}, {and} \bibinfo{person}{Carsten Eickhoff}.}
  \bibinfo{year}{2021}\natexlab{}.
\newblock \showarticletitle{A transformer-based framework for multivariate time
  series representation learning}. In \bibinfo{booktitle}{\emph{Proceedings of
  the 27th ACM SIGKDD Conference on Knowledge Discovery \& Data Mining}}.
  \bibinfo{pages}{2114--2124}.
\newblock


\bibitem[Zhang et~al\mbox{.}(2019)]%
        {zhang2019estimating}
\bibfield{author}{\bibinfo{person}{Gemeng Zhang}, \bibinfo{person}{Biao Cai},
  \bibinfo{person}{Aiying Zhang}, \bibinfo{person}{Julia~M Stephen},
  \bibinfo{person}{Tony~W Wilson}, \bibinfo{person}{Vince~D Calhoun}, {and}
  \bibinfo{person}{Yu-Ping Wang}.} \bibinfo{year}{2019}\natexlab{}.
\newblock \showarticletitle{Estimating dynamic functional brain connectivity
  with a sparse hidden Markov model}.
\newblock \bibinfo{journal}{\emph{IEEE Transactions on Medical Imaging}}
  \bibinfo{volume}{39}, \bibinfo{number}{2} (\bibinfo{year}{2019}),
  \bibinfo{pages}{488--498}.
\newblock


\end{thebibliography}

\clearpage
\appendix

\section{Models}
\subsection{Choice of temporal layers}
\begin{figure}[h!]
    \centering
    \captionsetup{font=small}
    \includegraphics[width=0.5\textwidth]{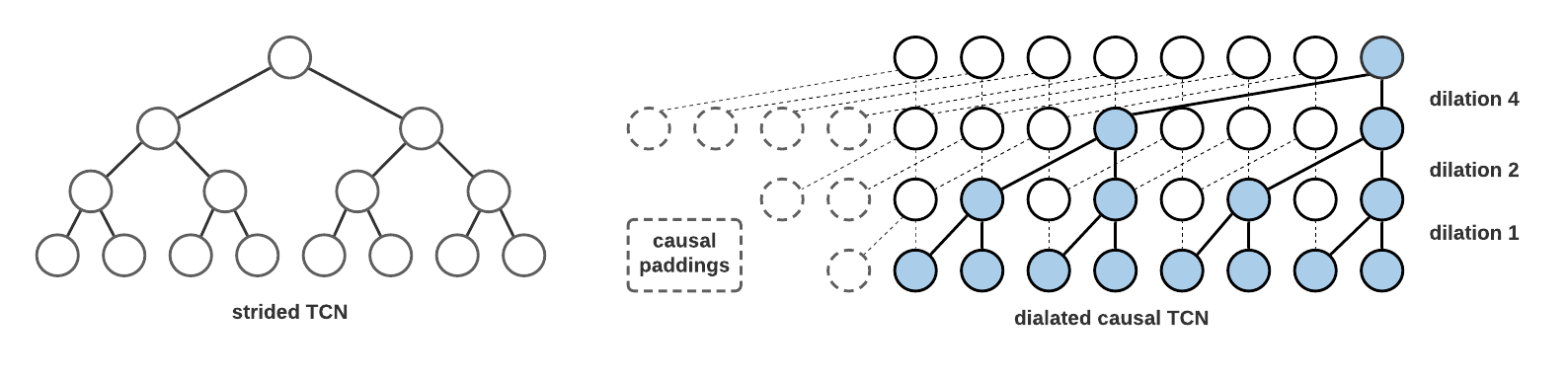}
    \vspace{-0.5cm}
    \caption{Comparison of strided non-causal TCN (left) and dilated causal TCN (right). For a causal TCN, the causal aspect is achieved through padding $(\text{kernel\_size} - 1) \times \text{dilation}$ number of zeros to the layer's input. The resulting $\mathbf{y}$ always has the same length as input $\mathbf{x}$, in which $\mathbf{y}_\tau$ only depends on inputs $\mathbf{x}_{t\leq\tau}$. We can view strided non-causal TCN as the rightmost node of a dilated causal TCN.}
    \label{fig:tcn}
    \vspace{-0.5cm}
\end{figure}

Fig. \ref{fig:tcn} explains the choice of TCN layers.

\subsection{Regularization terms for soft-assignment}
\label{ssc:pool_reg}
For each soft assignment matrix $S \in \mathbb{R}^{N \times c \times t}$ in \cref{eq:diffpool}, we test three regularization terms:
\begin{itemize}[leftmargin=*]
    \item Similar to \textsc{DiffPool}, to ensure a more clearly defined node assignment, namely each node is only assigned to few clusters (the closer to one the better), we minimize the entropy of single node assignments: $L_{E_1} = \frac{1}{c} \sum_{i=1}^{c}H(S_{i})$.
    \item To ensure a representation separation among nodes, meaning the assignment should not assign all the nodes a same way, we maximize the entropy of node assignment \textit{patterns} across all nodes: $L_{E_2} = - \frac{1}{c} \sum_{i=1}^{c}H(\sum_{j=1}^{n}S_{ij})$.
    \item To make the assignment along temporal axis smoother, we penalize assignment variances within a small time window $[\hat{t}, \hat{t}+\tau]$: $L_T = \frac{1}{t-\tau} \sum_{\hat{t} = 0}^{t-\tau}\sigma(S_{[\hat{t}, \hat{t}+\tau]})$, where $\sigma$ represents standard deviation.
\end{itemize}
Together with cross entropy classification loss $L_{CE}$, the final loss function of the model becomes:
\begin{equation}
    L_{reg} = \alpha_1 L_{CE} + \alpha_2 L_{E_1} + \alpha_3 L_{E_2} + \alpha_4 L_T,\quad \sum_i \alpha_i = 1
\end{equation}

\section{Experiments}

\subsection{Ablation studies}
\label{sssc:ab_study}

Numerical values of \cref{fig:graph_supports} are reported in \cref{tab:ab_study}. Training time ranges from 51 seconds / epoch for length-8 inputs to 298 seconds / epoch for length-256 inputs. Models converges to a relatively stable loss level within 20 epochs.

\begin{table}[h]
    \centering
    \captionsetup{font=small}
    \caption{Weighted F1 of ablation study settings.}
    \vspace{-0.2cm}
    \label{tab:ab_study}
    \resizebox{0.5\textwidth}{!}{%
    \begin{tabular}{c|cccccc}
    \toprule
    Input length (frames)  & 8      & 16     & 32     & 64     & 128    & 256    \\ \hline
    (i): SC + adp          & \textbf{66.19} & \textbf{70.18} & \textbf{75.87} & \textbf{76.14} & \textbf{82.91} & \textbf{90.85} \\
    (ii): SC only          & 64.54 & 65.58 & 71.79 & 70.31 & 73.63 & 89.79 \\
    (iii): adp only        & 64.32 & 65.20 & 74.01 & 71.42 & 80.63 & 89.46 \\
    (iv): SC + FC          & 66.10 & 67.58 & 70.26 & 75.02 & 76.91 & 84.68 \\
    (v): random adj   & 62.17 & 66.25 & 72.30 & 73.72 & 76.58 & 89.22 \\\hline
    (vi): (i) without smoothing & 63.57 & 62.82 & 70.19 & 65.82 & 72.91 & 79.65 \\
    (vii): (v) without smoothing & 56.88 & 64.08 & 72.27 & 62.72 & 75.16 & 83.75 \\
    (viii): coarsened graph & 37.92 & 42.23 & 46.18 & 52.12 & 57.17 & 64.25
    \\\bottomrule
    \end{tabular}}
\end{table}

\textbf{(I) Number of GNN layers.}
The total number of temporal layers depends on the input signal length since each strided $\operatorname{TCN}$ layer reduces the temporal length by a factor of two: if the input length is $2^i$, there need to be $i$ temporal layers. \textit{But is alternating every TCN with GNN the best strategy, or do we only need to follow one GNN after a few TCNs}? We study this question with different input lengths. 

\begin{figure}
    \centering
    \captionsetup{font=small}
        \includegraphics[trim={0.65cm 0 1cm 0.5cm}, clip, width=0.25\textwidth]{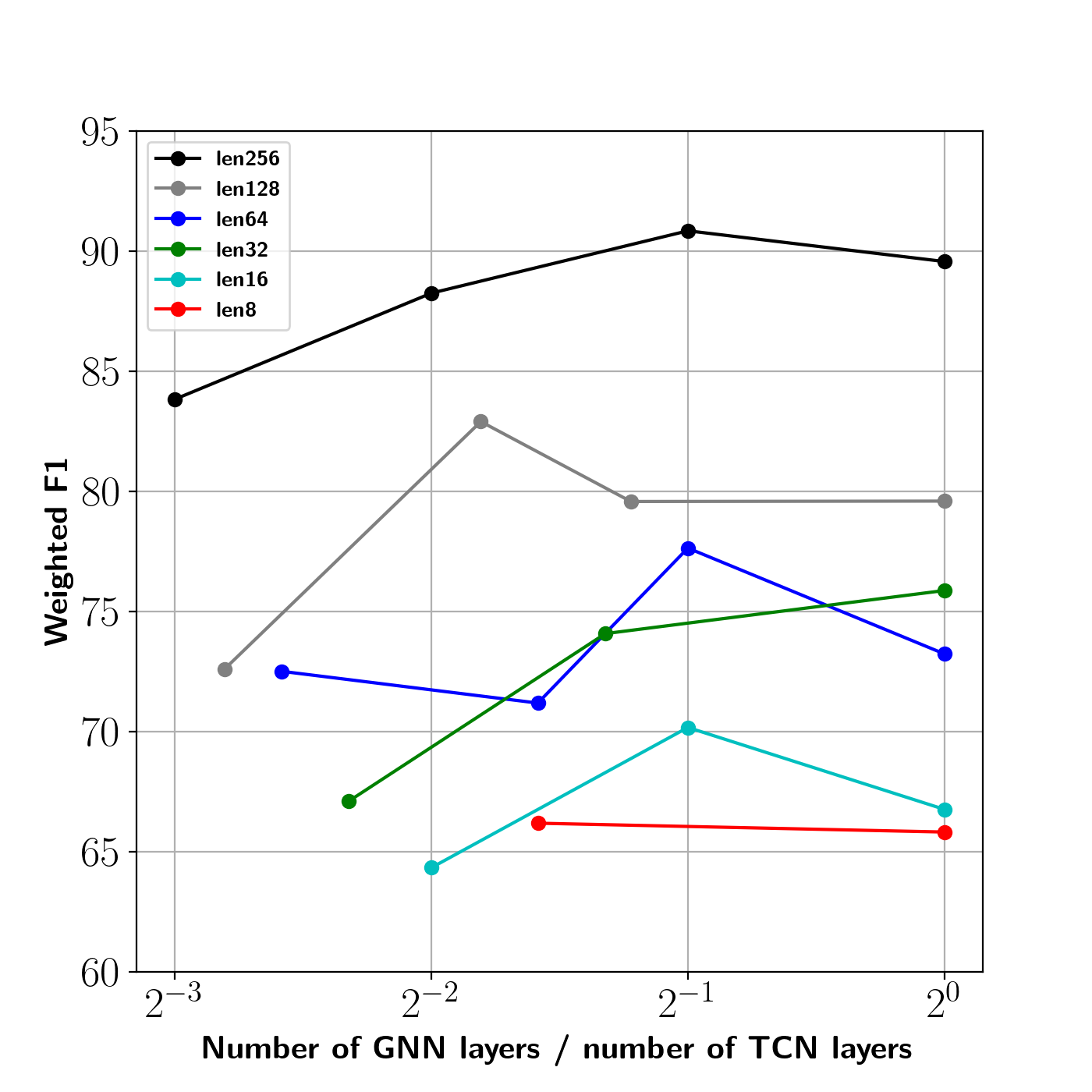}
    \vspace{-0.2cm}
    \caption{Choosing number of GNN to TCN layer ratio for different input lengths. In most cases, two TCN layers per GNN layer results in the best model performance in terms of F1.}
    \label{fig:number_gnn}
    \vspace{-0.5cm}
\end{figure}

Model weighted F1 are plotted in \cref{fig:number_gnn} for all possible $\operatorname{GNN}$ to total $\operatorname{TCN}$ ratios (e.g. length-256 inputs requires 8 $\operatorname{TCN}$ layers. The possible ratios are $\frac{1}{8}, \frac{1}{4}, \frac{1}{2}, 1$ since we can insert one $\operatorname{GNN}$ per 8, 4, 2, 1 $\operatorname{TCN}$ layers). The figure shows alternating every layer rarely yields the highest performance and the best ratio lies around one $\operatorname{GNN}$ per two $\operatorname{TCN}$ layers for our dataset. We repeat the experiment for $K=1,3$ (in \cref{eq:gnn}) to rule out the possibility that this result is related to how many neighbors one $\operatorname{GNN}$ layer can reach; we find they have roughly the same pattern as the $K=2$ case. We hypothesize that a lower $\operatorname{GNN}$ to $\operatorname{TCN}$ ratio does not capture enough spatial context, while higher ones might be overfitting. We leave exploring the relationship between this ratio and the number of nodes $N$ to a future study.

The best $\operatorname{GNN}$ to $\operatorname{TCN}$ ratio also depends on whether model incorporates latent adjacency matrices or not: without $A_{\text{adp}}$, length-128 signals achieves its relative best (among all ratios) when having one $\operatorname{GNN}$ per two $\operatorname{TCN}$s, but it only needs one $\operatorname{GNN}$ per three $\operatorname{TCN}$s if using $A_{\text{adp}}$. This shows learning latent structures $A_{\text{adp}}$ not only improves overall model accuracy but can also reduce model parameters, thus complexity, in achieving better results.

\begin{figure}
    \centering
    \captionsetup{font=small}
    \begin{subfigure}[b]{0.236\textwidth}
    \includegraphics[trim={20pt 0 20pt 10pt}, clip, width=\textwidth]{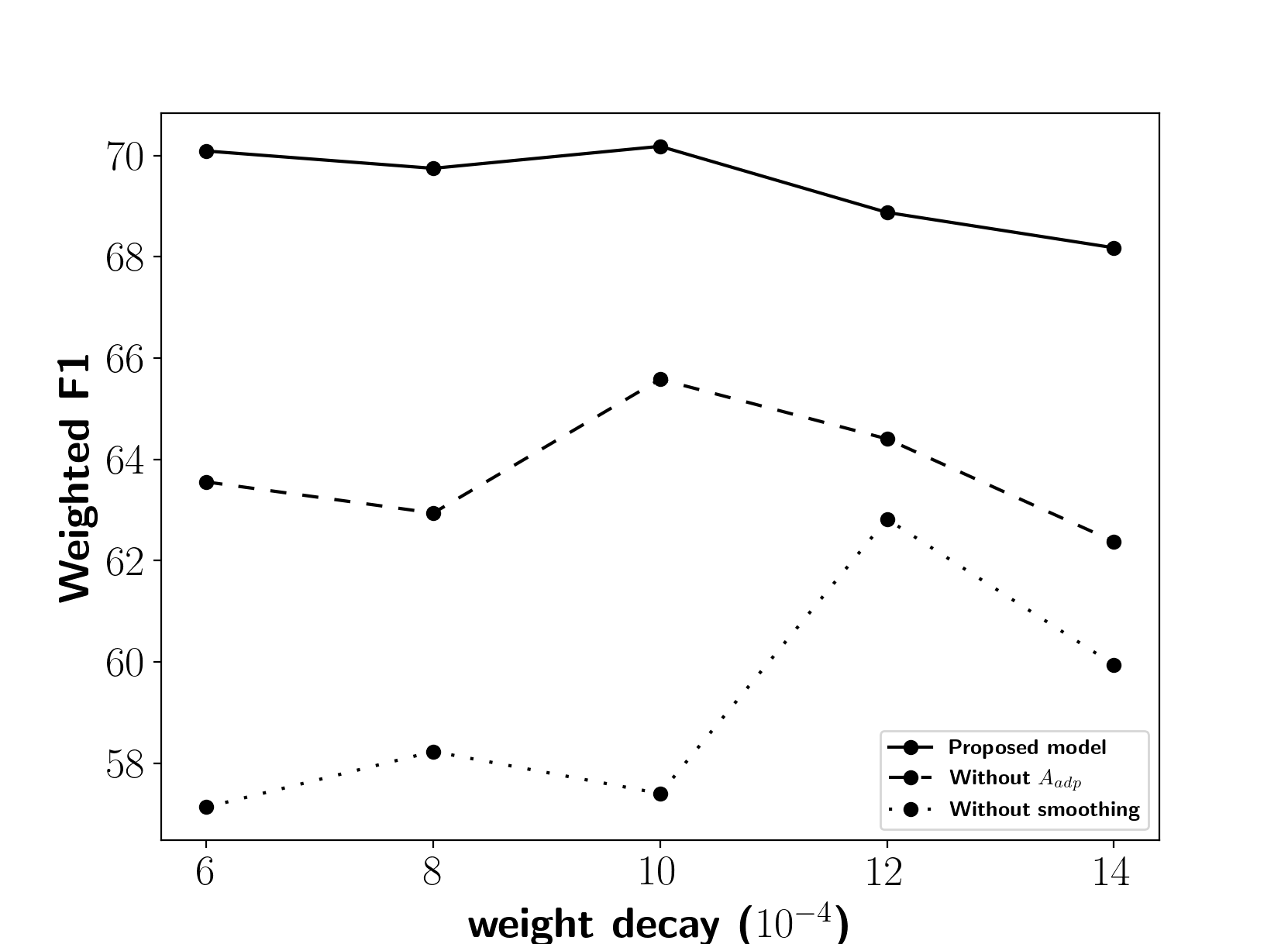}
    \caption{}
    \label{fig:grid_search}
    \end{subfigure}
    \begin{subfigure}[b]{0.236\textwidth}
    \includegraphics[trim={20pt 10pt 20pt 10pt}, clip, width=\textwidth]{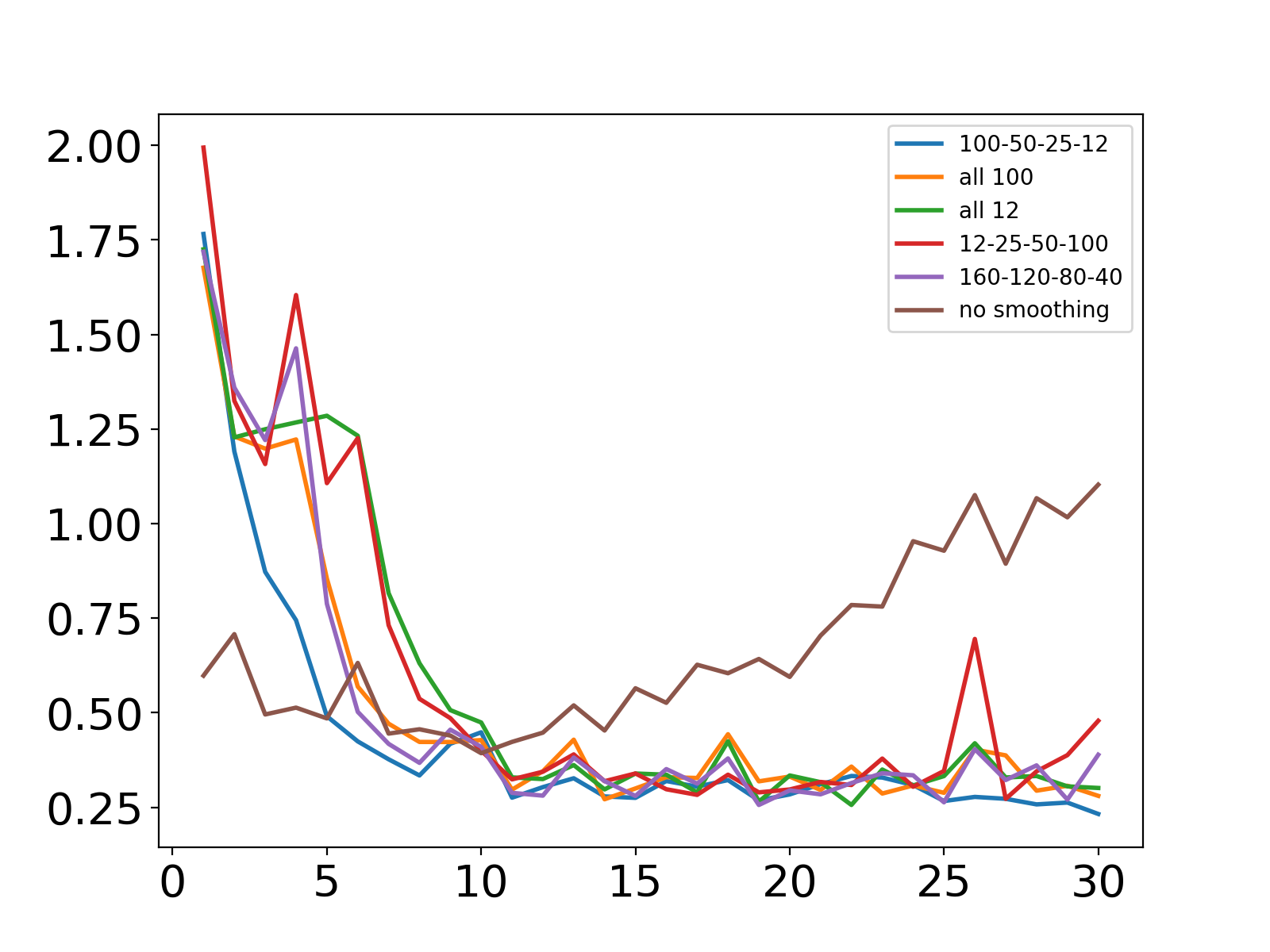}  
    \caption{}
    \label{fig:c_num}
    \end{subfigure}
    \vspace{-0.5cm}
    \caption{
    (a) adding inner cluster smoothing or input-dependent adaptive adjacency matrix makes the model more stable across various learning rates (results shown are from length-16 inputs).
    (b) Validation loss v.s. training epochs. Input length is 256, and four smoothing modules are used. Legends are the soft-assignment cluster numbers of the four smoothing modules. Our other experiments use decreasing cluster numbers that halved at each module, corresponding to the 100-50-25-12 choice here.}
    \vspace{-0.3cm}
    \label{fig:c_effect}
\end{figure}

\textbf{(II) Effects of soft-assignment cluster numbers.} During our experiments, we find that as long as the smoothing module is used, the final performance will be close to each other, only the convergence rates are different. Fig. \ref{fig:c_num} shows how validation loss converges with different $c$ (cluster number) or when there is no smoothing module. From it, we can observe that halving the numbers (100-50-25-12) is the most helpful setting, and we use it for our other experiments; decreasing the numbers (160-120-80-40) or all larger numbers (all 100) works better than increasing the numbers (12-25-50-100) or all smaller numbers (all 12). With the inner cluster smoothing module, all cluster number settings converge to around 0.23 at their smallest when trained for 30 epochs; their test weighted F1 range from 89.47 (model with 12-25-50-100) to 90.85 (model with 100-50-25-12).

On the contrary, if no smoothing module is used, the model overfits easily, and the validation loss can only reach about 0.4 before going up (with the best set of learning rate and weight decay parameters found with grid search). Understandably, the model is prone to overfitting given the complexity of $\operatorname{GNN}$ and the relatively small dataset size. However, our added inner cluster smoothing module effectively counters the effect and further brings the loss down in a stable manner. 

\begin{figure}
    \centering
    \captionsetup{font=small}
    \begin{subfigure}{0.38\textwidth}
    \includegraphics[width=\textwidth]{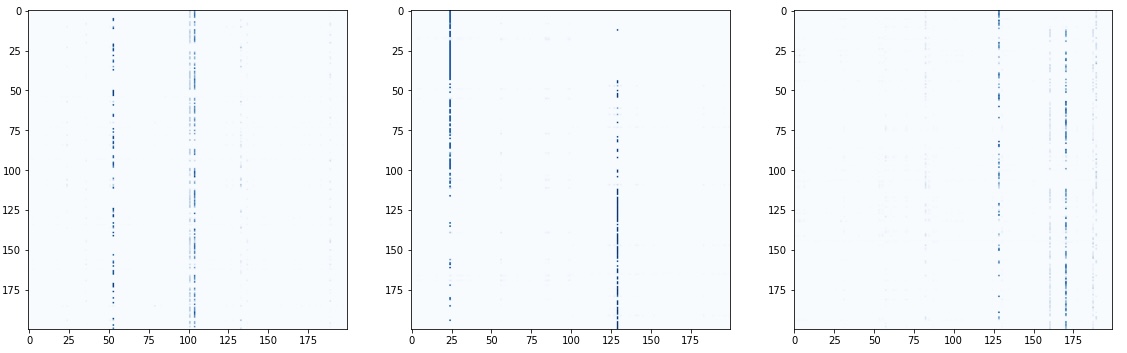}
    \caption{}
    \label{fig:adp_ind_task3}
    \end{subfigure}
    \begin{subfigure}{0.38\textwidth}
    \includegraphics[width=\textwidth]{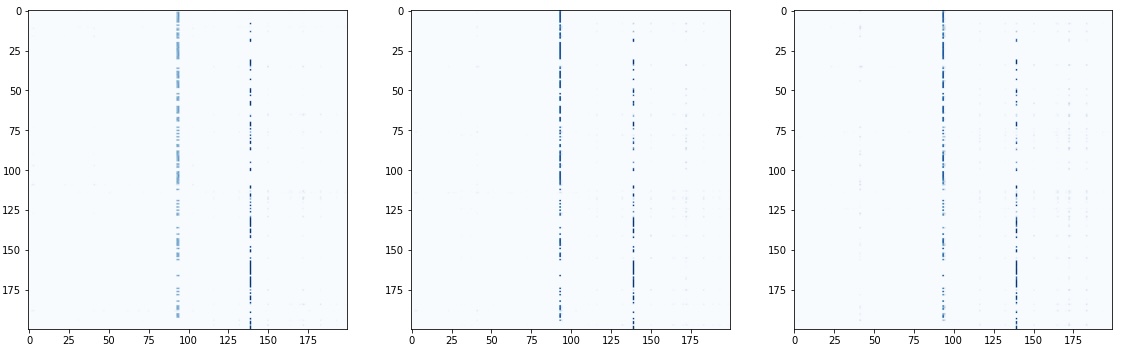}
    \caption{}
    \label{fig:adp_ind_task3_same_ses}
    \end{subfigure}    
    \begin{subfigure}{0.44\textwidth}
    \includegraphics[width=\textwidth]{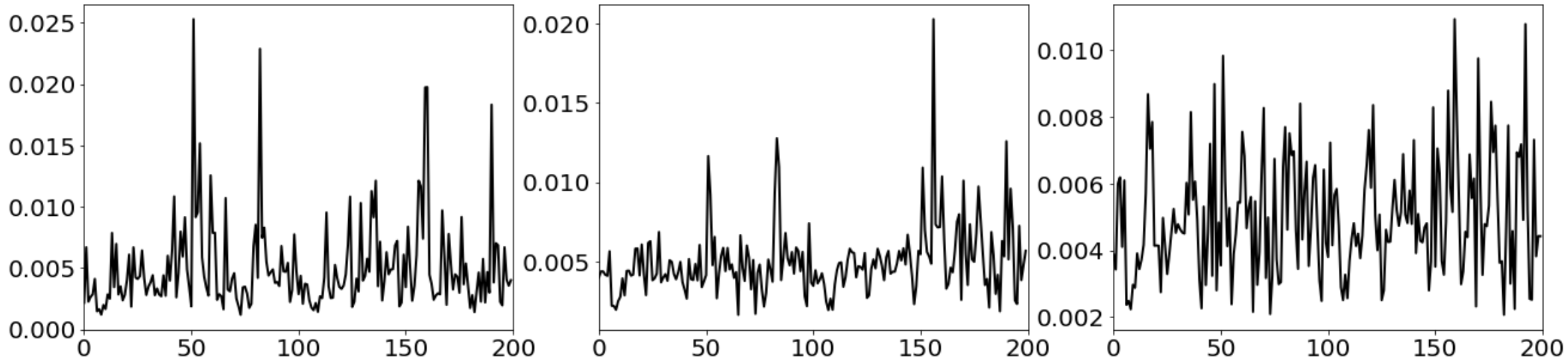}
    \includegraphics[width=\textwidth]{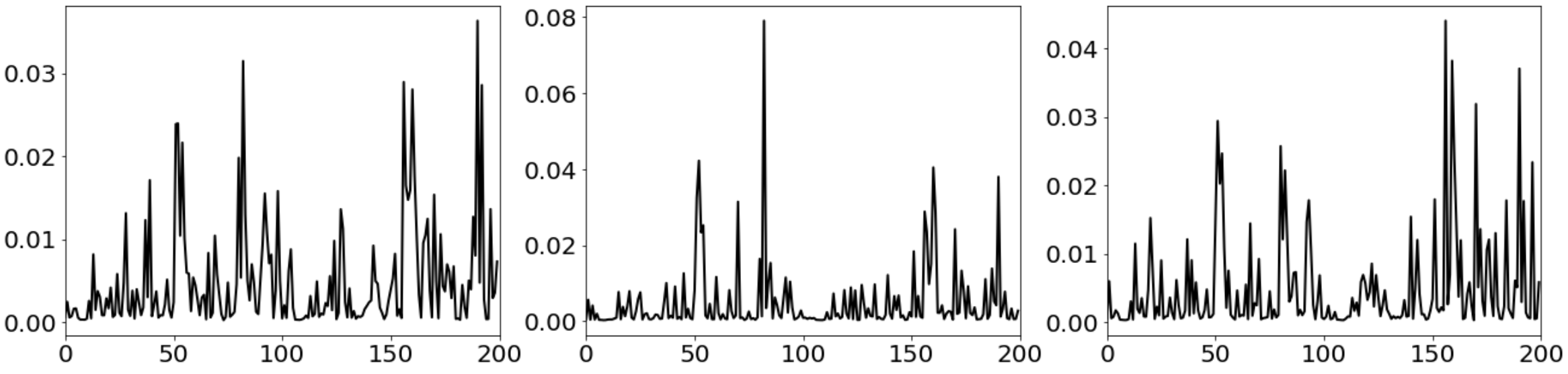}
    \caption{}
    \label{fig:adp_task_avg_roi}
    \end{subfigure}
    \begin{subfigure}{0.47\textwidth}
    \includegraphics[width=0.235\textwidth]{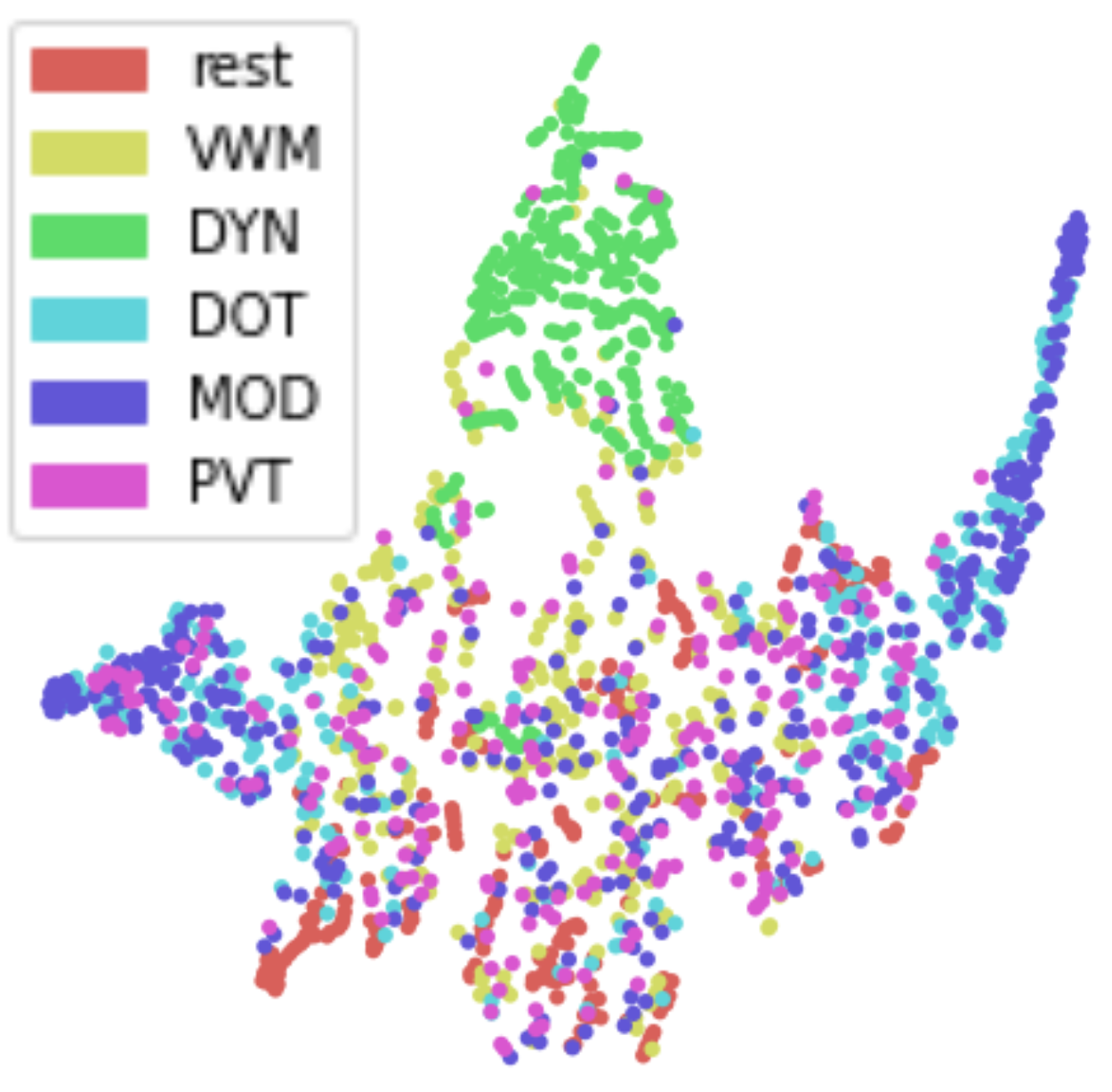}    
    \includegraphics[width=0.235\textwidth]{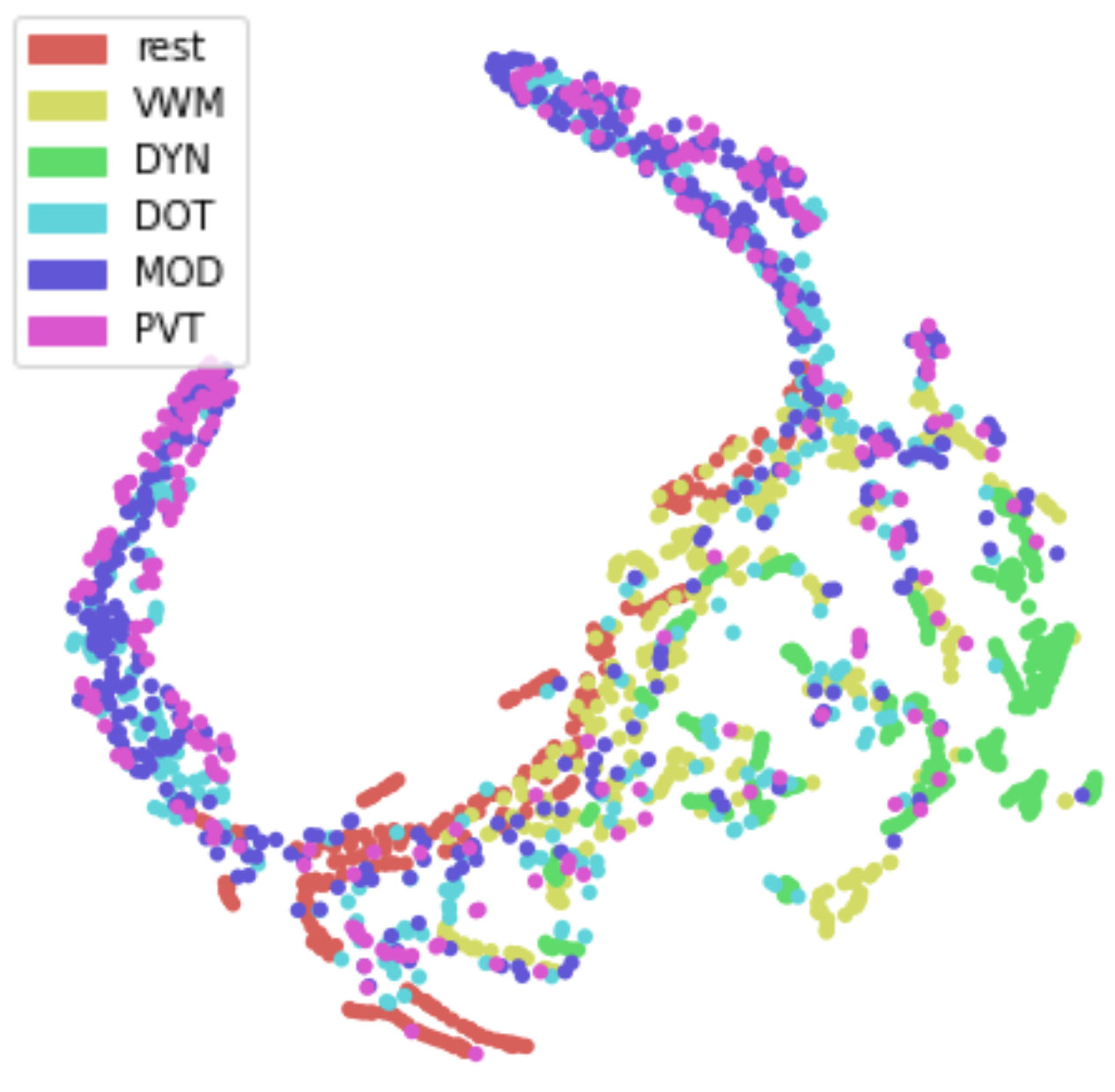}
    \includegraphics[width=0.235\textwidth]{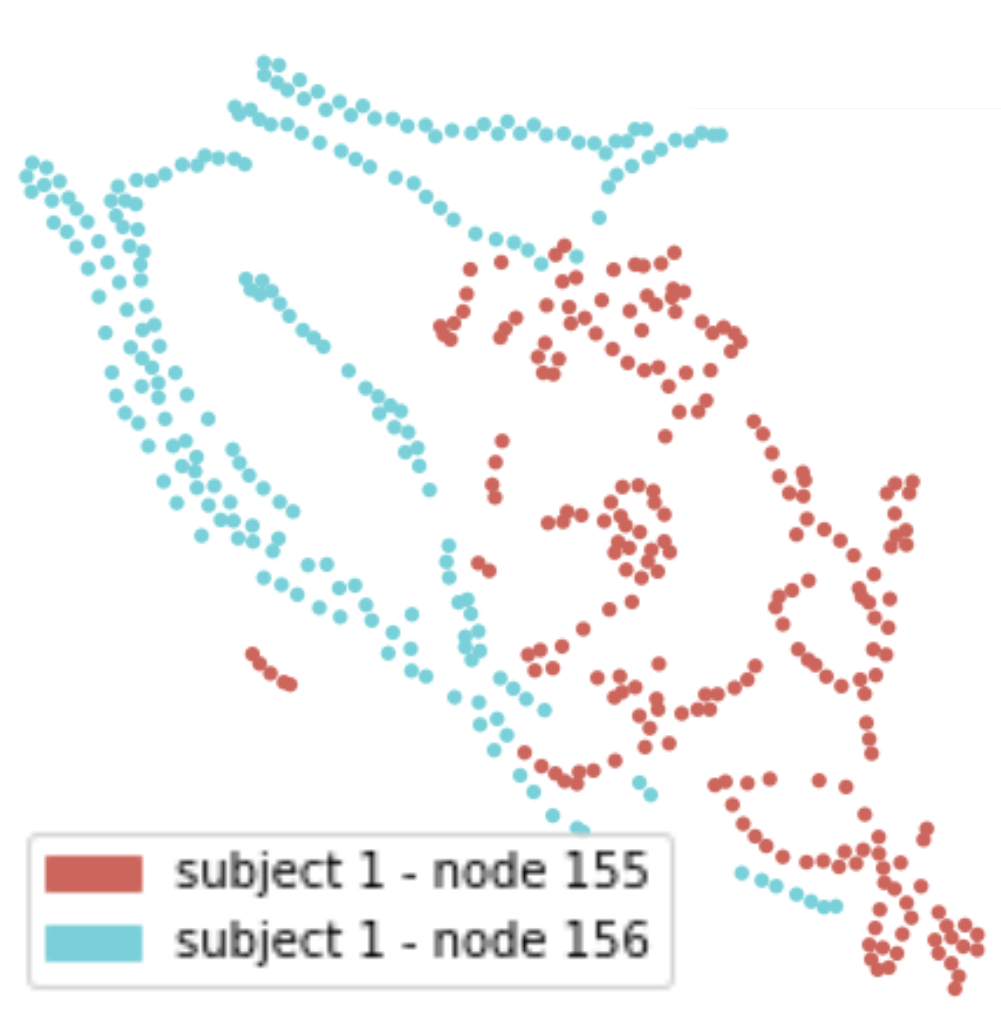}
    \includegraphics[trim={0 5pt 0 20pt}, clip, width=0.235\textwidth]{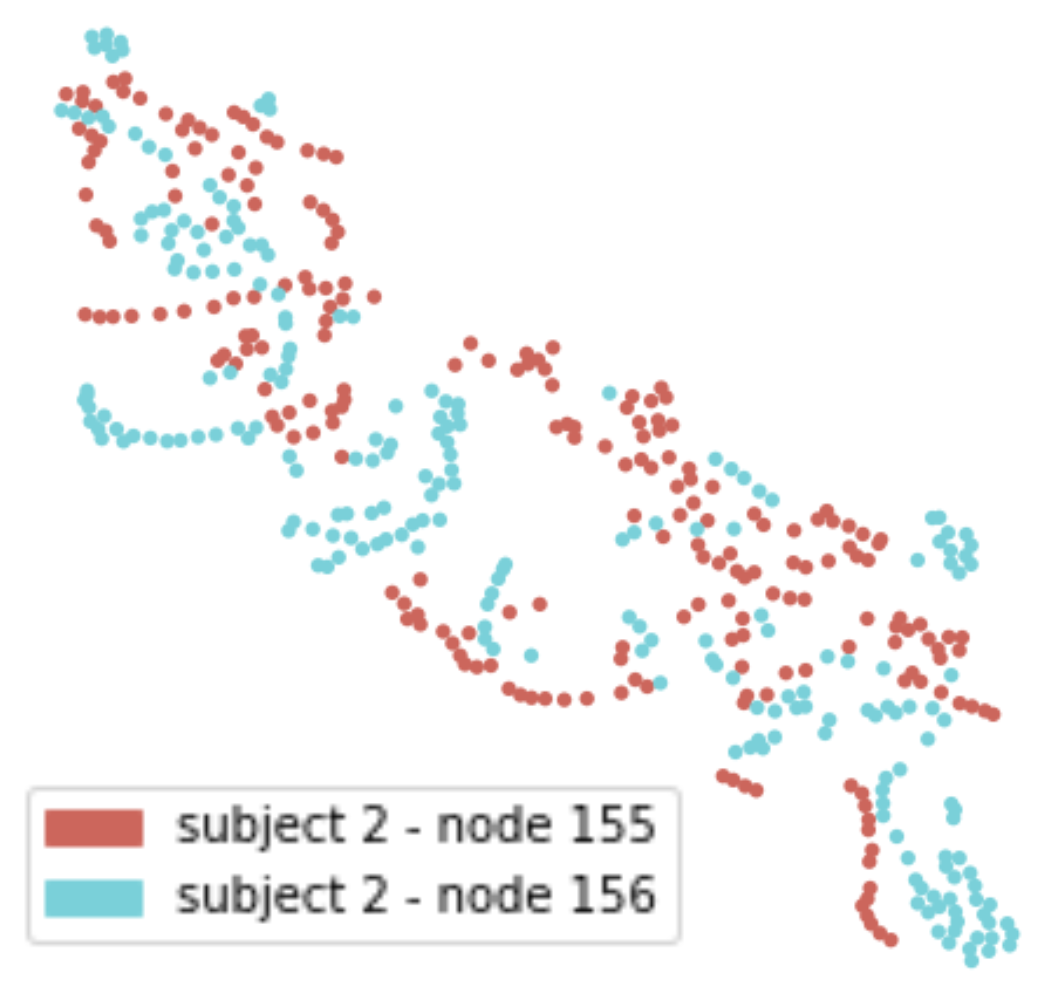}
    \caption{}
    \label{fig:nodevec}
    \end{subfigure}    
    \vspace{-0.3cm}
    \caption{Learned latent adaptive adjacency matrices.
    (a) $A_{i\_\text{adp}}$ of 3 randomly sampled inputs during the DOT task.
    (b) $A_{i\_\text{adp}}$ of 3 consecutive inputs from a same session during the DOT task. 
    (c) column averages of task-averaged $A_{\text{adp}}$ for resting state, VWM, DYN, DOT, MOD, PVT. 
    (d) left two: t-SNE of $X^{(\text{node-2, 156})} \Theta_{\text {adp }}$ in six tasks of one subject; right two: t-SNE of $X^{(\text{node-155, 156})} \Theta_{\text {adp }}$  during the resting state of two subjects (multiple sessions are aggregated).}
    \label{fig:adp}
\end{figure}

\subsection{Model comparisons}
\begin{figure}[t]
    \centering
    \captionsetup{font=small}
    \begin{subfigure}[b]{0.156\textwidth}
    \includegraphics[width=\textwidth]{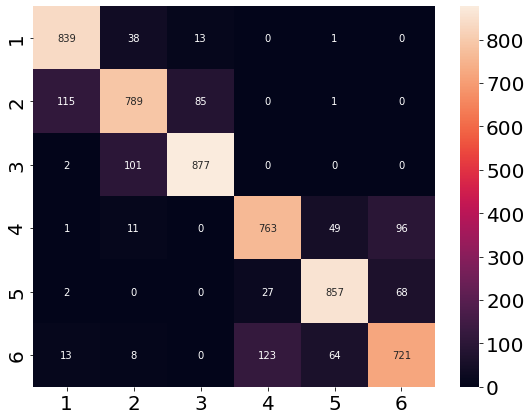}
    \caption{}
    \label{fig:cm_256}
    \end{subfigure}
    \begin{subfigure}[b]{0.156\textwidth}
    \includegraphics[width=\textwidth]{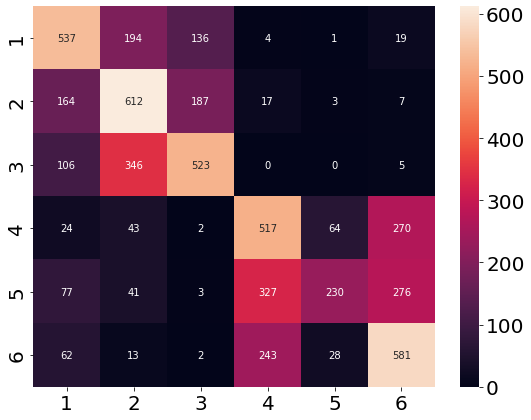}
    \caption{}
    \label{fig:cm_coarsened}
    \end{subfigure}
    \begin{subfigure}[b]{0.156\textwidth}
    \includegraphics[width=\textwidth]{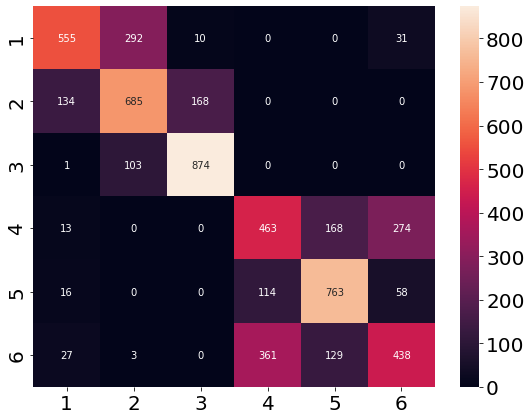}
    \caption{}
    \label{fig:cm_TG}
    \end{subfigure}
    \caption{Confusion matrices of:
    (a) ReBraiD (our proposed model),
    (b) model with coarsened graph (setting (viii)),
    (c) Graph Transformer (best graph baseline). Tasks are 1-Rest, 2-VWM, 3-DYN, 4-DOT, 5-MOD, 6-PVT.}
    \vspace{-0.2cm}
    \label{fig:cm}
\end{figure}

We plot confusion matrices of ReBraiD, the model from ablation study setting (viii), and the best performing graph baseline in \cref{fig:cm}. Misclassification pairs clustered at the first three tasks (resting, VWM, DYN) and the latter three (DOT, MOD, PVT). Shown confusion matrices are from models trained on length-256 inputs. We note that these misclassification pairs may differ for models trained on other input lengths (like 128-frame, etc.).

\subsection{Attributions}
\label{sssc:interp}
Many discriminatory regions obtained from $\operatorname{Attr}_A$ are consistent with existing literature:\\
\textbf{Resting state}: The top attributed ROIs belong to the default mode network, which is regarded salient during the resting state \citep{raichle2015brain}.\\
\textbf{VWM}: The dominant attributions are from visual regions and posterior parietal regions, which complies with \cite{todd2004capacity}.\\
\textbf{DYN}: Attributions from our model suggest regions along cingulate gyrus (defaultA-SalValAttnB-ContA-ContC-defaultC), as well as peripheral visual and somatomotor regions. Literature suggests anterior cingulate cortex (ACC) to be active \citep{kim2016anterior} and posterior cingulate cortex (PCC) to be inactive \citep{leech2014role} during visual attention tasks. This means both regions provide discriminative information about the DYN states, which is what our attribution method votes for.\\
\textbf{DOT}: Important ROIs from our analysis are located in control networks, in particular both ACC and PCC, as well as in the peripheral visual system. In the literature, dorsal and rostral regions of the ACC are proved to be involved with dot-probe performance \citep{carlson2012nonconscious, carlson2013functional}.\\
\textbf{MOD}: Our important ROIs are mostly in temporal-parietal regions and default mode network (anatomically frontoparietal), and literature suggests similar regions: parietal \citep{grabner2011brain} and prefrontal \citep{friedrich2013mathematical}.\\
\textbf{PVT}: Our top attributed ROIs belong to control networks, attention networks, and somatomotor regions. This is similar to \cite{drummond2005neural}, where both attention and motor systems are considered important.

\subsection{Inner cluster smoothing toy example.}
\label{ssc:toy_example}

Here we show a toy example demonstrating the inner cluster smoothing module described in \cref{eq:diffpool,eq:smoothing}. Note that we will only show one time slice, and the same operation is done along every $t$: on a particular $t$, we have $Z\in\mathbb{R}^{N\times d}, S\in\mathbb{R}^{N\times c}$. We will use $N=3, c=2$ and node values $a,b,c\in \mathbb{R}^d$ for this toy example.
In addition, this example is just to illustrate the concept behind the smoothing operation, and $\operatorname{Softmax}$ along the axis 1 is simplified as row normalization for a more straightforward presentation.

\begin{figure} [h]
    \centering
    \begin{minipage}[t]{0.4\textwidth}
    \includegraphics[trim={0 0 0 3.3cm}, clip, width=0.4\textwidth]{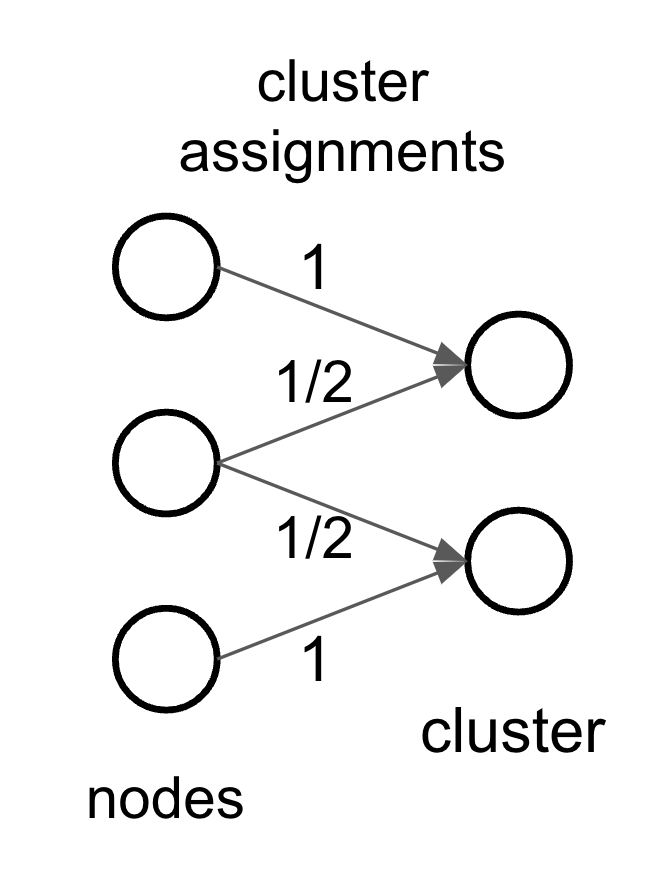}
    \end{minipage}
    \begin{minipage}[t]{0.1\textwidth}
      \vspace{-2.4cm}\(\text{\quad cluster assignment}\)
    \end{minipage}
    \begin{minipage}[t]{0.38\textwidth}
    \vspace{-0.7cm}
    \[
    \begin{aligned}
    &Z=\left(\begin{array}{l}
    a \\
    b \\
    c
    \end{array}\right), S=\left(\begin{array}{ll}
    1 & 0 \\
    \frac{1}{2} & \frac{1}{2} \\
    0 & 1
    \end{array}\right) \Rightarrow \tilde{H}=S^{\top} Z=\left(\begin{array}{l}
    a+\frac{1}{2} b \\
    \frac{1}{2} b+c
    \end{array}\right) \\
    &\tilde{S}=\text { row-normalized }\left(S^{T}\right)=\left(\begin{array}{ccc}
    \frac{2}{3} & \frac{1}{3} & 0 \\
    0 & \frac{1}{3} & \frac{2}{3}
    \end{array}\right) \\
    & \Rightarrow H_{new} = \tilde{S}^{\top} H=\left(\begin{array}{l}
    \frac{2}{3} a+\frac{1}{3} b \\
    \frac{1}{3} a+\frac{1}{3} b+\frac{1}{3} c \\
    \frac{1}{3} b+\frac{2}{3} c
    \end{array}\right)
    \end{aligned}
    \]
    \end{minipage}
    \caption{Inner cluster smoothing toy example.}
    \label{fig:toy_example}
\end{figure}

In this example, $1^{st}$ and $2^{nd}$ nodes are assigned to the first cluster, and $2^{nd}$ and $3^{rd}$ node are assigned to the second cluster. The final $H_{new}$ after our smoothing module will mingle the first two nodes' values, and the last two nodes' values (based on assignment weights) while keeping their original node number unchanged.

\subsection{Task descriptions.}
\label{ssc:task_des}
The following are task descriptions of CRASH (Cognitive Resilience and Sleep History) dataset:

\textbf{Resting state}: The subject simply lays in the scanner awake, with eyes open for 5 minutes.

\textbf{Visual working memory task (VWM)}: The subject is presented with a pattern of colored squares on a computer screen for a very brief period (100ms). After ~1000ms, they are presented with a single square and must determine if it is the same or different color as the previously presented square at that location. Responses are made with a button press (\cite{luck1997capacity}).

\textbf{Dynamic Attention Task (DYN)}: Two streams of orientation gratings are presented to the left and right of fixation.  Subjects monitor specified stream for a target (about 2 degree shift in orientation, clockwise or counter clockwise) that indicates whether the subject should continue to monitor the current stream (hold) or monitor the other stream (shift) and respond with a button press (\cite{yantis2002transient}).

\textbf{Dot Probe Task (Faces) (DOT)}: On each trial, two faces are presented, one neutral and the other happy or angry for 500ms. Then, either of two simple symbols is presented at the position of either of the faces. The subject must make a forced choice discrimination of the symbol. Reaction time differences as a function of the valance for the preceding facial expression are calculated.  There is increased variability of the bias with PTSD and fatigue (\cite{sipos2014postdeployment}).

\textbf{Math task (MOD)}: Subjects perform a modular math computation every 8 seconds and respond with a yes or no button press. The object of modular arithmetic is to judge the validity of problems such as 51=19(mod 4). One way to solve it is to subtract the middle number from the first number (i.e., 51–19) and then divide this difference is by the last number (32/4). If the dividend is a whole number, the answer is “true.” Otherwise the answer is false (\cite{mattarella2011choke}).

\textbf{Psychomotor vigilance task (PVT)}: The subject monitors the outline of a red circle on a computer screen for 10 minutes, and whenever a counter clockwise red sweep begins, they press a button as fast as possible.  Subjects are provided with response time feedback.  The experimenter records response latencies (\cite{loh2004validity}).

\end{document}